\documentclass[acmsmall, screen]{acmart}

\PassOptionsToPackage{table}{xcolor}
\usepackage[utf8]{inputenc} 
\usepackage[T1]{fontenc}    
\usepackage{hyperref}       
\usepackage{url}            
\usepackage{booktabs}       
\usepackage{amsfonts}       
\usepackage{nicefrac}       
\usepackage{microtype}      

\usepackage{multirow}
\usepackage{pifont}
\usepackage{amsthm}
\usepackage{graphicx}
\usepackage{amsmath}
\usepackage{algorithm}
\usepackage{algorithmic}
\usepackage{longtable}
\usepackage{adjustbox}
\usepackage[normalem]{ulem}
\useunder{\uline}{\ul}{}
\usepackage{wrapfig}
\usepackage{float}
\usepackage{subcaption}
\usepackage{enumitem}
\usepackage[utf8]{inputenc}
\usepackage{tcolorbox}
\tcbuselibrary{breakable}  

\AtBeginDocument{%
  }

\begin{document}

\title{RoboInspector: Unveiling the Unreliability of Policy Code for LLM-enabled Robotic Manipulation}

\author{Chenduo Ying}
\email{cdying@zju.edu.cn}
\orcid{0000-0001-8980-2314}
\affiliation{%
  \institution{Zhejiang University}
  \city{Hangzhou}
  \country{China}
}

\author{Linkang Du}
\orcid{0009-0004-9028-9326}
\affiliation{%
  \institution{Xi'an Jiaotong University}
  \city{Xi'an}
  \country{China}
  }
\email{linkangd@xjtu.edu.cn}

\author{Peng Cheng}
\orcid{0000-0002-4221-2162}
\affiliation{%
  \institution{Zhejiang University}
  \city{Hangzhou}
  \country{China}
}
\email{lunarheart@zju.edu.cn}

\author{Yuanchao Shu}
\authornote{Corresponding author.}
\orcid{0000-0002-9542-7095}
\affiliation{%
  \institution{Zhejiang University}
  \city{Hangzhou}
  \country{China}
}
\email{ycshu@zju.edu.cn}

\renewcommand{\shortauthors}{C. Ying et al.}

\begin{abstract}
  Large language models (LLMs) demonstrate remarkable capabilities in reasoning and code generation, enabling robotic manipulation to be initiated with just a single instruction. 
   The LLM carries out various tasks by generating policy code required to control the robot. 
   Despite advances in LLMs, achieving reliable policy code generation remains a significant challenge due to the diverse requirements of real-world tasks and the inherent complexity of user instructions.
   In practice, different users may provide distinct instructions to drive the robot for the same task, which may cause the unreliability of policy code generation. 
   To bridge this gap, we design \textbf{RoboInspector}, a pipeline to unveil and characterize the unreliability of the policy code for LLM-enabled robotic manipulation from two perspectives: the complexity of the manipulation task and the granularity of the instruction.
   We perform comprehensive experiments with 216 distinct combinations of tasks, instructions, and LLMs in two prominent frameworks.
   The \textbf{RoboInspector} identifies four main unreliable behaviors that lead to manipulation failure.
   We provide a detailed characterization of these behaviors and their underlying causes, giving insight for practical development to reduce unreliability.
   Furthermore, we introduce a refinement approach guided by failure policy code feedback that improves the reliability of policy code generation by up to 35\% in LLM-enabled robotic manipulation, evaluated in both simulation and real-world environments.
\end{abstract}




\maketitle

\section{Introduction}

\begin{wrapfigure}{r}{0.42\textwidth}  
  \centering
  \includegraphics[width=\linewidth]{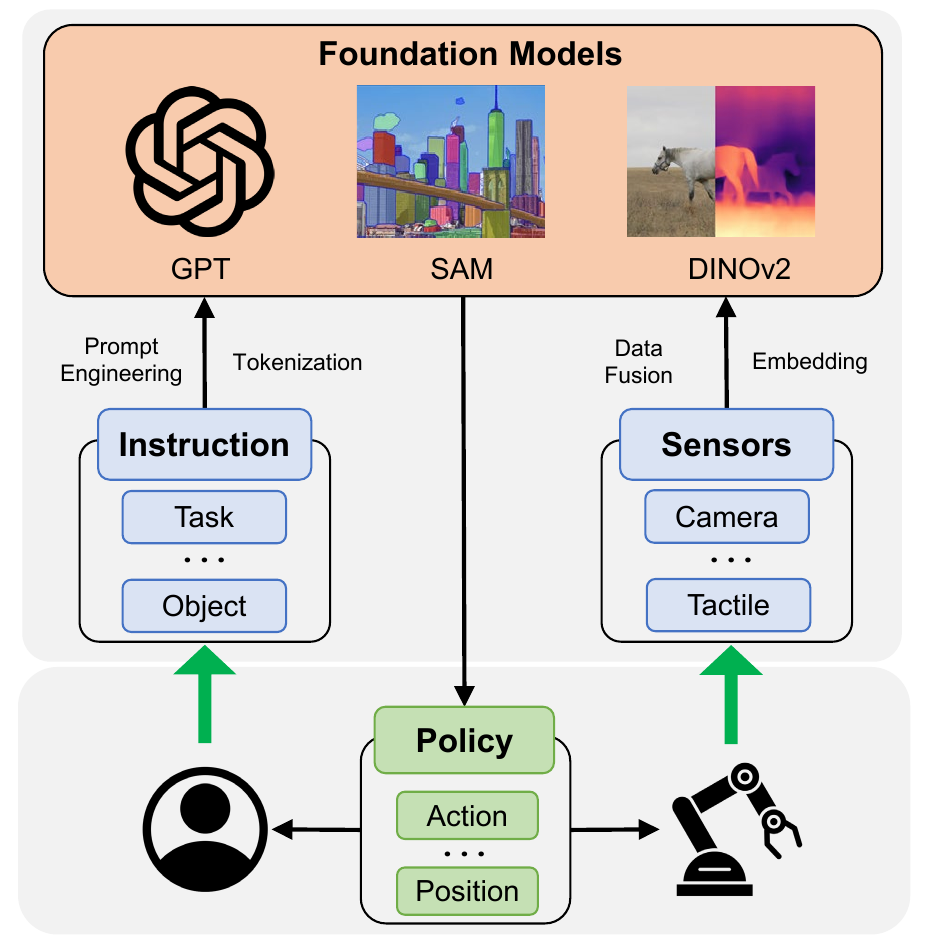}
  \caption{Illustration of LLM-enabled robotic manipulation general framework.}
  \Description{This figure shows an illustration of LLM-enabled robotic manipulation general framework.} 
  \label{fig:framework}
\end{wrapfigure}

The remarkable capabilities of foundation models, like large language models (LLMs) and vision-language models (VLMs), are widely recognized. Integrating these foundation models with embodied agents, such as robotic arms~\cite{brohan2023rt}, quadruped robots~\cite{hoeller2024anymal}, and humanoid robots~\cite{darvish2023teleoperation}, simplifies robotic manipulation to the level of requiring only one instruction as input.
As shown in \autoref{fig:framework}, the embodied agent processes user instructions and sensor inputs via foundation models~\cite{achiam2023gpt,kirillov2023segment,oquab2023dinov2} to generate policy outputs. 
Embodied agents are capable of independently sensing, learning, and interacting with their environments, which evolve consistently through dynamic interaction with the surroundings.

Embodied agents generally use LLMs and VLMs as their cognitive core to interpret instructions and generate policy codes to execute tasks. 
However, real-world instructions and tasks are intrinsically complex and varied. 
Specifically, different users employ varied natural language expressions to give instructions, and the necessary actions for the agent span a wide range of tasks.
These challenges may lead to substantial variations in the policy code generated by LLMs, even under identical environments. This raises concerns about the reliability of LLM-generated policy code in robotic manipulation.
The existing work~\cite{wu2024safety,zhang2024badrobot,liu2024exploring,chen2024towards} mainly focuses on the risks posed by adversarial attacks, such as BADROBOT~\cite{zhang2024badrobot} for jailbreaking embodied agents in the physical world. 
Nonetheless, these works neither consider benign instruction variations nor explore the impact of task complexity on the reliability of robotic manipulation.

To this end, we design \textbf{RoboInspector}, a general pipeline for unveiling the unreliability of LLM-generated policy code and characterizing four unreliable behaviors in robotic manipulation. 
Specifically, the pipeline focuses on the unreliability of LLM-generated policy code along two perspectives: the complexity of manipulation task and the granularity of user instructions. 
We conduct extensive trials using three series with a total of eight mainstream closed- and open-source LLMs. 
These trials unveil the impact of variations in the above two perspectives on the unreliability of LLM-generated policy code.
Based on the results, we characterize four types of unreliable behaviors that contribute to manipulation failure.
In addition, motivated by the insights gained from characterizing unreliable behaviors, we present a feedback-based refinement approach and demonstrate its effectiveness through experiments.

In summary, our contributions are three-fold:
\begin{itemize}[leftmargin=20pt]
    \item By decomposing the primitive actions involved in common robotic manipulation tasks and analyzing the compositional structure of instructions, along with extensive experiments, we unveil that the unreliability of LLM-generated policy code in mainstream frameworks is primarily influenced by two factors: task complexity and instruction granularity.
    \item We design \textbf{RoboInspector} and perform comprehensive experiments with 216 distinct combinations of tasks, instructions, and LLMs. Based on the review of generated policy code, we identify four types of unreliable behaviors—\textit{Nonsense}, \textit{Disorder}, \textit{Infeasible}, and \textit{Badpose}—that lead to manipulation failure. We further provide a detailed performance characterization of these behaviors and give insight of cause. 
    \item We introduce a refinement approach guided by failure policy code and unreliable behaviors description feedback. Our method achieves up to a 35\% improvement in manipulation success rate across both simulated and real-world settings.
\end{itemize}

\section{Related Works}

\paragraph{LLM-enabled Robotic Manipulation.} The integration of pre-trained language models into embodied agents is a dynamic research field~\cite{vemprala2024chatgpt,yang2023auto,10.1145/3719664}. The researchers aim to improve planning and reasoning capabilities~\cite{huang2022language,ahn2022can,zeng2022socratic}. Recent works use LLMs for contextual grounding in interactive planning scenarios. Frameworks such as React~\cite{yao2022react} and ECoT~\cite{zawalski2024robotic} incorporate chain-of-thought reasoning~\cite{wei2022chain}. These methods enable agents to develop autonomous problem-solving policies. In robotics, various methods help bridge the perception-action gap. Examples include providing descriptions of textual scenes~\cite{zeng2022socratic,huang2022inner,singh2023progprompt}, using perception APIs~\cite{liang2023code}, and incorporating vision during decoding~\cite{huang2023grounded} or as input through multimodal models~\cite{driess2023palm}.
Additionally, work~\cite{meng2025embodied} proposed a few-shot learning approach that combines progressive structural exemplars with Chain-of-Thought-guided LLMs to generate robust and generalizable policy code for robotic manipulation.
LLMs demonstrate behavioral commonsense for basic control tasks. However, their ability to perform spatial-level composition remains uncertain. Predefined motion primitives are still necessary for effective action execution. Research on robot reward specification is another focus, the literature explores reward design~\cite{kwon2023reward}, exploration strategies~\cite{colas2020language}, and preference learning~\cite{hu2023language}. LLM-generated rewards have been tested using physics simulators like MuJoCo~\cite{yu2023language}. Advancements in foundation models further improve robotic manipulation. Examples include SAM~\cite{kirillov2023segment}, DINOv2~\cite{oquab2023dinov2}, and SORA~\cite{brooks2024video}.
In addition to the related works mentioned above, the survey~\cite{yao2023bridging} provides a comprehensive discussion of various studies on language instructions in robotic manipulation, including their cognitive roles.
Despite these developments, research on robotic manipulation in terms of safety and reliability is still lacking.

\paragraph{Robotic Manipulation Safety and Reliability.} The safety and reliability risks associated with robotic manipulation are a critical yet underexplored area of research. Recent studies have highlighted issues such as the generation of offensive language~\cite{azeem2024llm}, potential privacy breaches, and the dissemination of misinformation~\cite{hundt2022robots} by LLM-enabled embodied agents. Furthermore, the susceptibility of LLMs to adversarial attacks is a focal point of research. Some works~\cite{wu2024safety,liu2024exploring,mahiul2024malicious} explored how subtle modifications to input can cause LLMs to produce incorrect or harmful outputs. These vulnerabilities are particularly concerning in applications where LLMs are integrated into robotic manipulation decision-making processes. Additionally, the threat of backdoor attacks is a concern~\cite{liu2024compromising}. The potential for backdoor attacks underscores the necessity for rigorous testing and validation of LLMs before their deployment in embodied agents. In response to these challenges, comprehensive safety frameworks~\cite{zhang2024safeembodai,zhu2024eairiskbench} have been proposed to guide the responsible development and deployment of the embodied agent. These frameworks advocate for continuous evaluation and adaptation to mitigate emerging risks associated with robotic manipulation. While significant advancements have been made in safety risks, there remains a pressing need to address the reliability challenges of robotic manipulation. This paper aims to unveil and characterize the unreliability of policy code in LLM-enabled robotic manipulation.

\section{RoboInspector}

\begin{figure*}[!t]
      \centering
      \includegraphics[width=\linewidth]{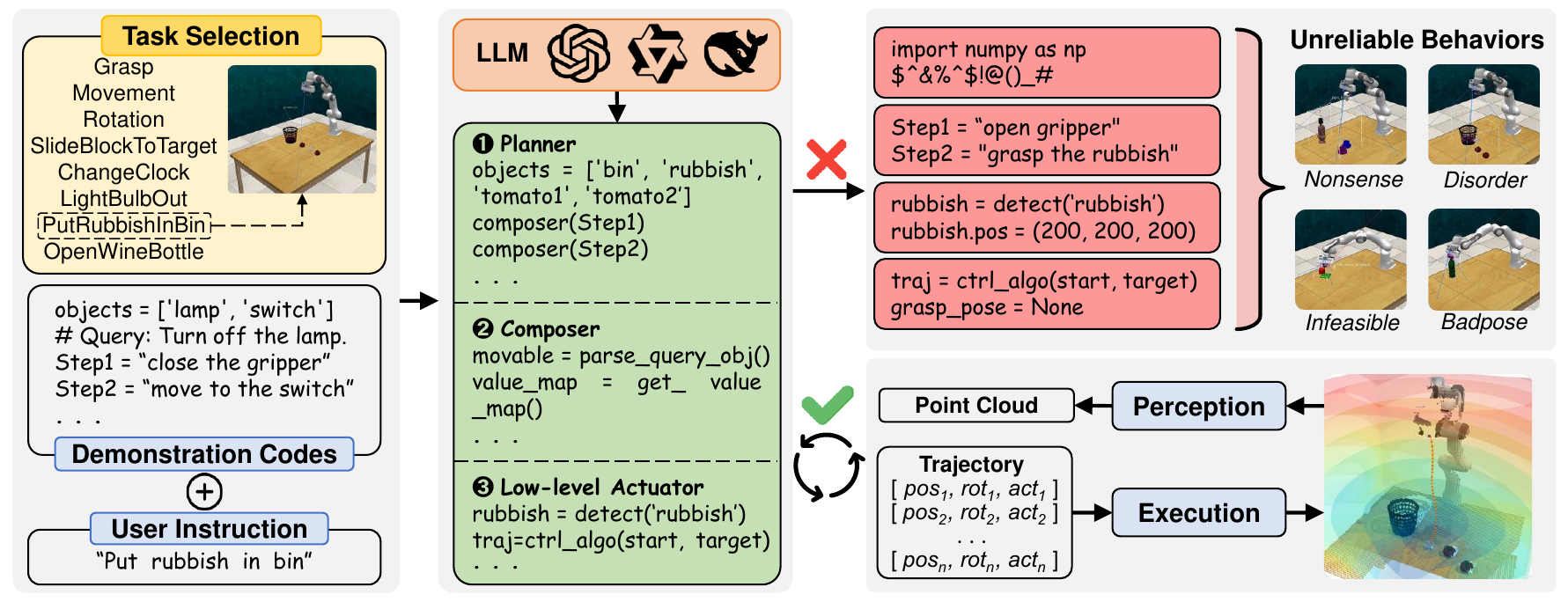}
      \caption{\textbf{RoboInspector Pipeline.} User instruction can be adjusted to requirements. Green block show correct LLM-generated cascaded codes. Red blocks indicate unreliable ones.}
      \Description{This figure shows the pipeline for unveiling the unreliability of LLM-enabled robotic manipulation.} 
      \label{fig:pipeline}
\end{figure*} 

\subsection{Overview}
Methods for achieving robotic manipulation include task planning~\cite{yao2022react,ahn2022can}, reward learning~\cite{kim2024openvla,bharadhwaj2024roboagent}, policy code generation~\cite{huang2023voxposer,liang2023code,ma2023eureka}, etc.
Among them, policy code generation is characterized by low design complexity and high solution efficiency, making it a promising way for LLM-enabled robotic manipulation. 

In view of the above situation, \textbf{RoboInspector} focuses mainly on the generating of policy codes, as shown in~\autoref{fig:pipeline}.
The pipeline constructs 216 distinct combinations by selecting different tasks, instructions, and LLMs.
By analyzing and decomposing tasks and instructions, we categorize common robotic manipulation tasks into three levels of task complexity and three levels of instruction granularity.
The manipulation success rate serves as a metric to reflect the impact of task complexity and instruction granularity on the policy code generating reliability for robotic manipulation.
We take VoxPoser~\cite{huang2023voxposer} and Code as Policies~\cite{liang2023code} as examples.
Note that the policy code generation frameworks of LLM-enabled robotic manipulation are similar.
Thus, the \textbf{RoboInspector} can be modified for use in other systems based on LLM-generated policy code.
In general, the workflow begins with the environment settings, where users give instructions based on the task and their requirements. 
The LLM interprets the instructions to generate policy code that directs the embodied agent in performing primitive actions to complete the manipulation.
In the following sections, we provide a detailed description of each module within the \textbf{RoboInspector}.


\subsection{Task Selection}
\begin{wraptable}{r}{0.45\textwidth} 
\centering
\caption{Primitive actions involved in various tasks.}
    \label{tab:1}
    \small  
    \begin{tabular}{@{}cccc@{}}
    \toprule
                                & \multicolumn{3}{c}{\textbf{Primitive Action}}                                                          \\ \cmidrule(l){2-4} 
\multirow{-2}{*}{\textbf{Task}} & \textit{Grasp}                            & \textit{Move}                             & \textit{Rotate}                           \\ \midrule
\textbf{Grasp}                           & {\color[HTML]{32CB00} \ding{52}} & {\color[HTML]{FE0000} \ding{56}} & {\color[HTML]{FE0000} \ding{56}} \\
\textbf{Movement}                        & {\color[HTML]{FE0000} \ding{56}} & {\color[HTML]{32CB00} \ding{52}} & {\color[HTML]{FE0000} \ding{56}} \\
\textbf{Rotation}                        & {\color[HTML]{FE0000} \ding{56}} & {\color[HTML]{FE0000} \ding{56}} & {\color[HTML]{32CB00} \ding{52}} \\
\textbf{SlideBlockToTarget} & {\color[HTML]{FE0000} \ding{56}} & {\color[HTML]{32CB00} \ding{52}} & {\color[HTML]{FE0000} \ding{56}} \\
\textbf{ChangeClock}                     & {\color[HTML]{32CB00} \ding{52}} & {\color[HTML]{FE0000} \ding{56}} & {\color[HTML]{32CB00} \ding{52}} \\
\textbf{LightBulbOut}                    & {\color[HTML]{FE0000} \ding{56}} & {\color[HTML]{32CB00} \ding{52}} & {\color[HTML]{32CB00} \ding{52}} \\
\textbf{PutRubbishInBin}    & {\color[HTML]{32CB00} \ding{52}} & {\color[HTML]{32CB00} \ding{52}} & {\color[HTML]{FE0000} \ding{56}} \\
\textbf{OpenWineBottle}     & {\color[HTML]{32CB00} \ding{52}} & {\color[HTML]{32CB00} \ding{52}} & {\color[HTML]{32CB00} \ding{52}} \\ \bottomrule
\end{tabular}
\end{wraptable}
By analyzing and decomposing common robotic manipulations, we find that nearly all manipulations consist of three primitive actions: \textit{Grasp}, \textit{Move}, and \textit{Rotate}.
We denote the set of primitive actions as $\mathcal{S} =\{\textit{Grasp},\textit{Move},\textit{Rotate}\}$. First we develop three simple manipulation tasks focusing exclusively on the three primitive actions in the simulations of both VoxPoser and Code as Policies: \textbf{Grasp}, \textbf{Movement}, and \textbf{Rotation}.
Further we select five manipulation tasks inherently offered by RLBench~\cite{james2020rlbench}, with adjustments to adapt our experiment. Except for the \textbf{SlideBlockToTarget} task, all these five tasks involve two or more primitive actions. The \textbf{SlideBlockToTarget} task involves only the \textit{Move} action, providing a reference for comparison. The development scene is shown in Appendix \ref{sec:CaP-scene} and \ref{sec:VoxPoser-scene}. In addition, \autoref{tab:1} outlines the primitive actions involved in each task.
Each task $T_i$ ($i = 1, 2, \ldots, 8$), composed of different primitive actions, is selected in the pipeline to represent varying levels of manipulation complexity. The set of tasks is defined as $\mathcal{T} = \{T_1, T_2, \ldots, T_8\}$, where $T_i \subseteq \mathcal{S}$ and $T_i \neq \emptyset$. Note that $T_i$ denotes a manipulation task as an experimental unit, while the elements $\textit{Grasp},\textit{Move},\textit{Rotate} \in \mathcal{S}$ denote the primitive actions that constitute it. The complexity of executing a manipulation task is defined by the number of distinct primitive actions it contains, given as $f_\phi(T_i) := |T_i| \in \{1, 2, 3\}$, and increases with the number of these actions.

\subsection{Instruction Construction}
\label{sec:ins}
In \textbf{RoboInspector}, user instructions are combined with a piece of demonstration code to form a complete prompt, which is then fed into the LLM as input.
The demonstration code includes objects, query statements, and the corresponding action and code for these objects and queries.
Utilizing its in-context learning \cite{lin2023unlocking} and generalization abilities, the LLM produces outputs aligned with the user instruction and demonstration code.
The demonstration code is kept constant during the process to assess the performance of LLM-generated policy code across varying granularity levels of user instructions.
The granularity of user instructions refers to the quantitative relationship among the constituent elements included in an instruction statement.
For example, a user instruction $\mathcal{I}$ can be defined as a quadruple $\mathcal{I} := (\mathcal{O}, \mathcal{A}, \mathcal{P}, \mathcal{C})$, where $\mathcal{O}$, $\mathcal{A}$, $\mathcal{P}$, and $\mathcal{C}$ represent $object$, $action$, $purpose$, and $condition$, respectively. 
We define ${I_A} = \mathcal{O} \times \mathcal{A}$, representing an instruction statement that consists of only two elements: the object and the action (\emph{e.g.},``\textit{throw the rubbish}''). 
Similarly, ${I_P} = \mathcal{O} \times \mathcal{A} \times \mathcal{P}$ represents an instruction statement containing three elements: the object, action, and purpose (\emph{e.g.}, ``\textit{drop the rubbish into the bin}''). 
Lastly, ${I_C} = \mathcal{O} \times \mathcal{A} \times \mathcal{P} \times \mathcal{C}$ represents an instruction statement comprising the object, action, purpose, and condition. 
An example is ``\textit{Grasp the rubbish and place it in the bin, with the executable space defined as (100, 100, 100)}". 
Based on the above definitions, we create three types of instruction,${I_A^i}$, ${I_P^i}$, and ${I_C^i}$ for each task $T_i$. 
To denote the granularity of the instructions, the function $\boldsymbol{f}_\theta$ is defined as $\boldsymbol{f}_\theta(\mathcal{I}^i)= \mathcal{G}(\mathcal{I}^i)$, where 
\begin{equation}
    \mathcal{G}(\mathcal{I}^i) = \lvert \{ e \in \{O, A, P, C\} \mid e \neq \varnothing \} \rvert.
\end{equation}
It can be observed that the granularity of an instruction increases with the number of its constituent elements.

\subsection{Result Processing}
\label{sec:res}
After task selection and instruction construction, we begin the result processing.
On the basis of the received prompt, the LLM generates the corresponding policy code and actions.
Specifically, the model first generates a high-level planner that is responsible for understanding the instruction and decomposing the task into several specific steps for the composer to execute. 
Subsequently, for each step, the composer sequentially produces the execution code and identifies the low-level actuator necessary for execution.
Finally, the low-level actuator invokes the relevant API to complete the task.

There may be two situations.
In successful cases, the program outputs an optimized robot execution trajectory, which includes position, rotation angles, and gripper actions.
The robot control program performs manipulation through the execution module, and throughout the process, the perception module continuously collects sensor data, feeding it back to the LLM to confirm whether the manipulation is completed. 
However, the response from the LLM is unreliable as the large language model is inherently stochastic.
In \autoref{fig:pipeline}, we summarize four types of unreliable behavior and show the corresponding code representation for each case. 
A detailed case characterization will be provided in \autoref{sec:analysis}. 
Here, we express the success of the manipulation task execution as reliability $R$. 
The relevance between robotic manipulation's reliability, manipulation task complexity, and instruction granularity can be assumed as:
\begin{equation}
    R\big(\boldsymbol{f}_\phi(\mathcal{T}), \boldsymbol{f}_\theta(\mathcal{I})\big) 
    \propto \frac{\boldsymbol{f}_\theta(\mathcal{I}^i)}{\boldsymbol{f}_\phi(\mathcal{T}_i)}.
\end{equation}
The reliability value $R$ is positively correlated with $\boldsymbol{f}_\theta(\mathcal{I}^i)$ and negatively correlated with $\boldsymbol{f}_\phi(T_i)$.
We then conduct thorough experiments with \textbf{RoboInspector} to unveil the impact of instruction granularity and manipulation task complexity on policy code generation.

\section{Experiments}

\subsection{Settings}
\label{sec:setup}
\paragraph{Target LLMs.} We select seven commercial closed-source LLMs from two current mainstream series: the OpenAI GPT series~\cite{GPT2024} and Alibaba Cloud's Qwen series~\cite{Qwen2024}.
In addition, we chose a recently noteworthy LLM in the field of open-source LLM, DeepSeek-V3~\cite{DeepSeek2025} produced by DeepSeek Inc.
To make the model's response more definite while maintaining some variability, the `temperature' for chat completion is fixed at 0.1, with other parameters remaining at default settings.

\paragraph{Target Frameworks.} 
We take VoxPoser~\cite{huang2023voxposer} and Code as Policies~\cite{liang2023code} as prominent robotic manipulation frameworks to study unreliability of policy code generation.

\paragraph{Simulation.} 
We select the RLBench and PyBullet simulators as the platform for robotic manipulation.
They not only offer a variety of everyday tasks composed of various primitive actions, but also facilitates the creation of custom tasks. 
Each task scene in \autoref{tab:1} is created by randomly generating relevant objects within the workspace.
To ensure that the results are more generalizable, we made slight adjustments to these tasks, expanding the randomized workspace for generating scene objects. 
This adjustment better reflects the complexity of the real world.

\subsection{Main Results}
\label{sec:main}

We perform comprehensive experiments integrating manipulation tasks with differing complexity and user instructions of various granularity to unveil how these factors influence the reliability of policy code generation.
The experiment prompt template can be seen in Appendix \ref{sec:VoxPoser-prompt}.
Each task is tested 50 times under different instructions and LLMs, with a complete execution of the manipulation counted as a success. 
For example, in the \textbf{Grasp} task shown in \autoref{tab:2}, the manipulation success rate is 0.46 when using instruction ${I_A}$ and the GPT-3.5-turbo, indicating 23 complete and successful manipulation of 50 trials in VoxPoser framework.
The trials conducted in Code as Policies framework are presented in Appendix \ref{sec:CaP-res}.
The first three tasks only involve the execution of their respective primitive actions, and therefore the ${I_A}$ and ${I_P}$ instructions constructed for these tasks are identical.
To avoid redundancy, we only present the data for ${I_A}$ here.

From \autoref{tab:2}, we can summarize two main observations.
\textbf{1}) The three primitive action tasks exhibit a relatively high success rate in manipulation task. 
Even with the lowest granularity instructions ${I_A}$, each task achieves a success rate of approximately 70\%, which increases to around 82\% when using the highest granularity instructions ${I_C}$.
\textbf{2}) As the complexity of manipulation task increases, the average success rate decreases correspondingly. 
Moreover, for the same task, higher-granularity instructions consistently result in higher success rates.
This supports the notion that instruction granularity is positively correlated with higher success rates.
Compared to the three simple tasks mentioned above, all five complex tasks exhibit lower success rates. 
Notably, although the \textbf{SlideBlockToTarget} task involves only a single primitive action, its completion goal and environment are more complex, leading to a lower success rate.
In summary, the experimental results validate our assumption: manipulation task complexity is negatively correlated with reliability, while instruction granularity is positively correlated with reliability.
However, there are still a large number of failure cases waiting for us to explore here.

\begin{table}[t]
\centering
\caption{\textbf{Comparison of manipulation success rate on 168 distinct combinations of task, instruction, and LLM with VoxPoser framework (Code as Policies framework see Appendix \ref{sec:CaP-res} and \ref{sec:CaP-Statistics}). Detailed statistical data see Appendix \ref{sec:VoxPoser-statistics}.} Black bold underline marks the model with the highest success rate for same instruction, blue bold underline marks the instruction with the highest average success rate for same task. \textbf{Ins.} stands for Instructions.}
\label{tab:2}
\resizebox{\textwidth}{!}{%
\begin{tabular}{@{}cc|ccccccccc@{}}
\toprule
 &  & \multicolumn{7}{c}{\textbf{Closed-source LLM}} & \textbf{Open-source LLM} &  \\ \cmidrule(lr){3-10}
\multirow{-2}{*}{\textbf{Task}} & \multirow{-2}{*}{\textbf{Ins.}} & GPT-3.5-turbo & GPT-4 & GPT-4o & GPT-4o-mini & Qwen-max & Qwen-plus & Qwen-turbo & DeepSeek-V3 & \multirow{-2}{*}{\textbf{\thinspace Avg. \thinspace}} \\ \midrule
 & ${I_A}$ & 0.46 & 0.74 & 0.70 & 0.78 & 0.74 & 0.76 & 0.50 & {\ul \textbf{0.84}} & 0.69 \\
\multirow{-2}{*}{\textbf{Grasp}} & ${I_C}$ & 0.66 & 0.94 & 0.92 & 0.86 & 0.92 & 0.90 & 0.58 & {\ul \textbf{0.94}} & {\color[HTML]{3166FF} {\ul \textbf{0.84}}} \\ \midrule
 & ${I_A}$ & 0.58 & 0.80 & 0.78 & 0.74 & 0.78 & 0.74 & 0.58 & {\ul \textbf{0.80}} & 0.73 \\
\multirow{-2}{*}{\textbf{Movement}} & ${I_C}$ & 0.68 & 0.90 & 0.90 & 0.86 & {\ul \textbf{0.98}} & 0.86 & 0.54 & 0.90 & {\color[HTML]{3166FF} {\ul \textbf{0.83}}} \\ \midrule
 & ${I_A}$ & 0.50 & 0.70 & 0.74 & {\ul \textbf{0.76}} & 0.74 & 0.70 & 0.54 & 0.76 & 0.68 \\
\multirow{-2}{*}{\textbf{Rotation}} & ${I_C}$ & 0.64 & 0.90 & 0.86 & 0.88 & {\ul \textbf{0.90}} & 0.86 & 0.60 & 0.86 & {\color[HTML]{3166FF} {\ul \textbf{0.81}}} \\ \midrule
 & ${I_A}$ & 0.28 & 0.80 & 0.74 & 0.66 & 0.80 & 0.60 & 0.26 & {\ul \textbf{0.84}} & 0.62 \\
 & ${I_P}$ & 0.32 & 0.88 & 0.82 & 0.78 & {\ul \textbf{0.88}} & 0.72 & 0.38 & 0.86 & 0.71 \\
\multirow{-3}{*}{\textbf{SlideBlockToTarget}} & ${I_C}$ & 0.52 & 0.90 & 0.84 & 0.82 & {\ul \textbf{0.94}} & 0.80 & 0.44 & 0.94 & {\color[HTML]{3166FF} {\ul \textbf{0.78}}} \\ \midrule
 & ${I_A}$ & 0.20 & 0.72 & 0.62 & 0.70 & {\ul \textbf{0.74}} & 0.60 & 0.26 & 0.68 & 0.57 \\
 & ${I_P}$ & 0.24 & 0.78 & 0.82 & 0.74 & {\ul \textbf{0.82}} & 0.70 & 0.30 & 0.76 & 0.65 \\
\multirow{-3}{*}{\textbf{ChangeClock}} & ${I_C}$ & 0.44 & 0.84 & {\ul \textbf{0.88}} & 0.80 & 0.86 & 0.76 & 0.32 & 0.88 & {\color[HTML]{3166FF} {\ul \textbf{0.72}}} \\ \midrule
 & ${I_A}$ & 0.16 & 0.72 & {\ul \textbf{0.80}} & 0.70 & 0.74 & 0.60 & 0.18 & 0.72 & 0.58 \\
 & ${I_P}$ & 0.18 & 0.78 & {\ul \textbf{0.84}} & 0.76 & 0.80 & 0.62 & 0.22 & 0.74 & 0.62 \\
\multirow{-3}{*}{\textbf{LightBulbOut}} & ${I_C}$ & 0.32 & 0.84 & {\ul \textbf{0.90}} & 0.80 & 0.84 & 0.78 & 0.26 & 0.82 & {\color[HTML]{3166FF} {\ul \textbf{0.70}}} \\ \midrule
 & ${I_A}$ & 0.24 & 0.70 & 0.64 & 0.40 & 0.66 & 0.46 & 0.16 & {\ul \textbf{0.70}} & 0.50 \\
 & ${I_P}$ & 0.30 & {\ul \textbf{0.84}} & 0.82 & 0.50 & 0.72 & 0.50 & 0.20 & 0.80 & 0.59 \\
\multirow{-3}{*}{\textbf{PutRubbishInBin}} & ${I_C}$ & 0.40 & {\ul \textbf{0.92}} & 0.86 & 0.62 & 0.78 & 0.70 & 0.30 & 0.90 & {\color[HTML]{3166FF} {\ul \textbf{0.69}}} \\ \midrule
 & ${I_A}$ & 0.04 & 0.30 & 0.12 & 0.12 & 0.10 & 0.10 & 0.06 & {\ul \textbf{0.30}} & 0.14 \\
 & ${I_P}$ & 0.04 & 0.38 & 0.18 & 0.10 & 0.16 & 0.10 & 0.02 & {\ul \textbf{0.42}} & 0.18 \\
\multirow{-3}{*}{\textbf{OpenWineBottle}} & ${I_C}$ & 0.10 & {\ul \textbf{0.60}} & 0.30 & 0.18 & 0.32 & 0.14 & 0.10 & 0.52 & {\color[HTML]{3166FF} {\ul \textbf{0.27}}} \\ \bottomrule
\end{tabular}%
}
\end{table}

To provide insight into the computational cost associated with policy code generation, we also report the token usage for complete manipulation across different LLM series in \autoref{cost}. The complete token usage includes all API calls involved in one full manipulation attempt, covering the planner, composer, and low-level actuator stages. As shown in \autoref{tab:5}, a failed manipulation consumes significantly fewer tokens than a successful one on average. This discrepancy is expected, as failures typically occur early in the execution pipeline, resulting in fewer overall API interactions. In terms of latency, the GPT series and Qwen series exhibit moderate per-call response times of 3.26s and 2.67s respectively, while DeepSeek-V3 shows a higher average latency of 6.74s per call, likely attributable to server-side load during the experimental period.

\subsection{Further Analysis and Characterization of Unreliable Behaviors}
\label{sec:analysis}
\begin{figure*}[!t]
      \centering
      \includegraphics[width=\linewidth]{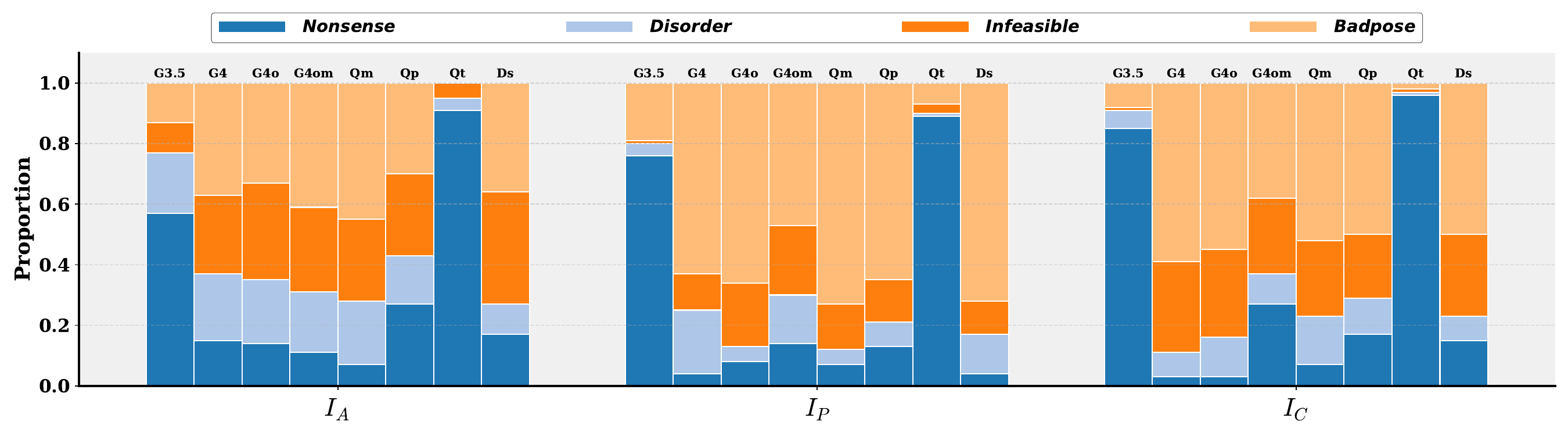}
      \caption{Proportion of unreliable behaviors contributing to manipulation failure for each model under different instructions.}
      \Description{This figure shows the proportion of unreliable behaviors contributing to manipulation failure for each model under different instructions.} 
      \label{fig:3}
\end{figure*}

\begin{figure*}[!t]
      \centering
      \includegraphics[scale=0.5]{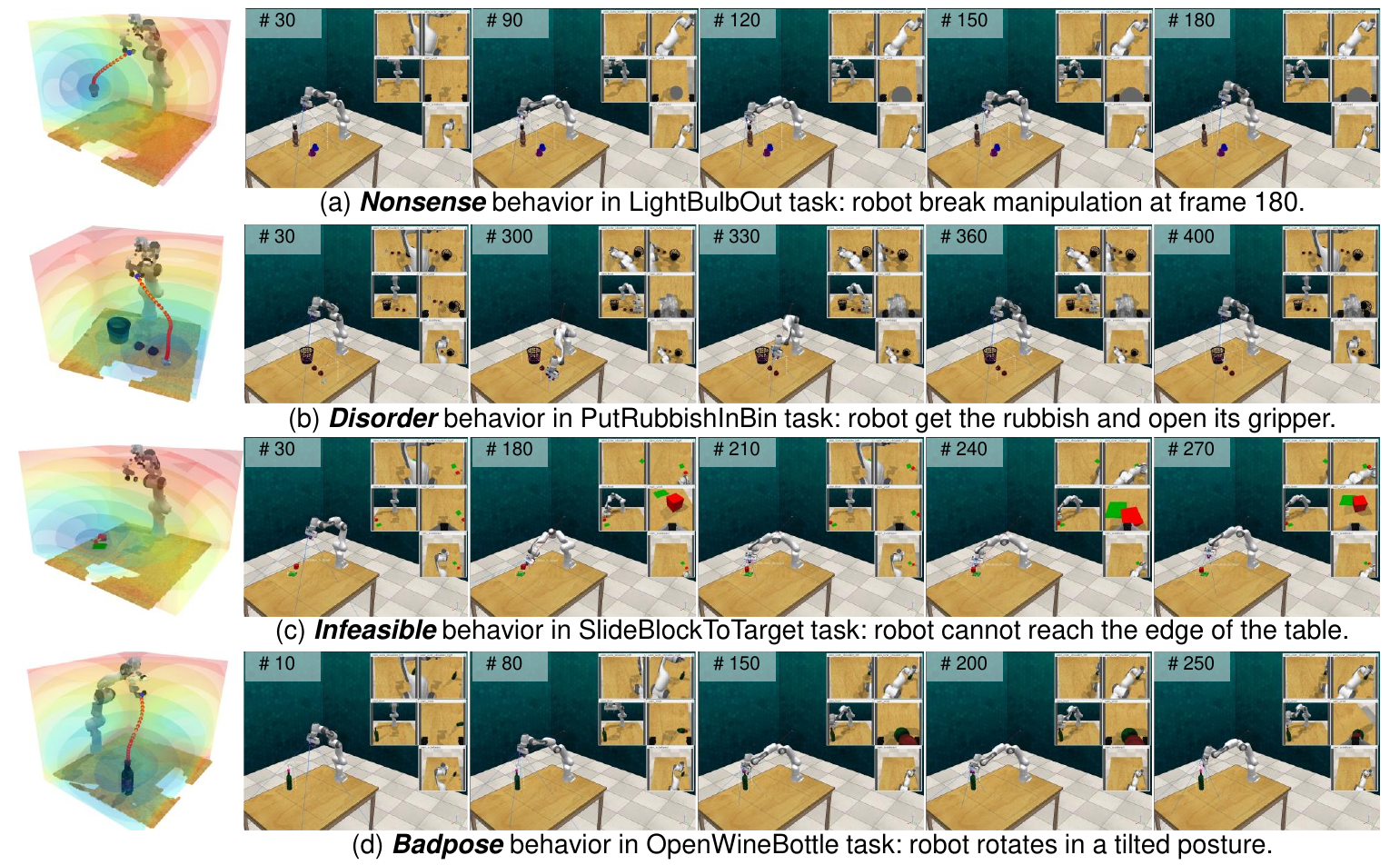}
      \caption{Examples of each unreliable behavior. The \# above the image represent frames.}
      \Description{This figure shows the examples of each unreliable behavior.} 
      \label{fig:4}
\end{figure*}

To provide insight into the failure manipulation cases, we analyze the policy code generated by LLMs and the environment in each experiment.
Overall, we identify four types of unreliable behaviors leading to the failure of LLM-enabled robotic manipulation.
In \autoref{fig:3}, we calculate the proportion of unreliable behaviors that contribute to manipulation failure for each model under different instructions.
Moreover, we provide examples of each unreliable behavior in \autoref{fig:4}.
Below, we delve into four types of unreliable behaviors.

\paragraph{\ding{182} \textbf{\textit{Nonsense}}.}
This describes cases where the LLM generates policy code that either does not conform to defined criteria or contains irrelevant text.

Recalling \autoref{sec:ins}, the LLM receives two components as input: the user instruction, selected according to the level of granularity, and the demonstration code, which assists the LLM in learning the desired output format via in-context learning.
On the one hand, the demonstration code includes the ``import" statements.
Before sending the constructed instructions to the LLM, we add several instruction prefixes. 
These prefixes include restrictive guidelines for LLMs, such as prohibiting the output of ``import" statements, as they interfere with the automated code execution. 
On the other hand, except for the query statements, all other parts of the demonstration code are Python code without any explanatory descriptions.
We expect the LLM to generate output consisting solely of executable Python code.

From \autoref{fig:3}, it is observed that the majority of failure cases for both GPT-3.5-turbo and Qwen-turbo are attributed to \textbf{\textit{Nonsense}} behavior, whereas other models exhibit a relatively lower proportion of such behavior. 
In this case, LLMs generates ``import" or redundant explanatory text, or a combination of both.
The LLM-generated policy code cannot be executed automatically, which causes the robot to cease all manipulations, as shown in \autoref{fig:4}a.
The first red block in \autoref{fig:pipeline} depicts the output of policy code with this unreliable behavior.

This revelation suggests that GPT-3.5-turbo and Qwen-turbo demonstrate weaker instruction-following capabilities. 
In the practical deployment of LLM-enabled robotic manipulation, it is advisable to avoid models with weaker performance to reduce the occurrence of \textbf{\textit{Nonsense}} behavior.


\paragraph{\ding{183} \textbf{\textit{Disorder}}.} It refers to the unreasonable sequence of manipulation steps generated by the LLM.

Recalling \autoref{sec:res}, the LLM generates the policy code and actions based on the given instructions.
Regardless of the task's complexity, the LLM first attempts to understand the instructions and decomposes the task objective into individual composers.
Each composer represents a substep necessary to accomplish the manipulation, executed sequentially in the prescribed order.
This implies that the order of the generated composers is crucial.
If the sequence is disordered, completing the manipulation becomes infeasible. 

Taking the task of \textbf{PutRubbishInBin} as an example, the correct sequence of composers should involve executing ``move to the bin" first, followed by ``open gripper". 
However, in cases of failure caused by \textbf{\textit{Disorder}} behavior, the LLM generates sequences such as opening the gripper before moving to the bin.
The robot follows unreasonable steps to perform manipulations and undoubtedly fails, as shown in \autoref{fig:4}b.
The second red block in \autoref{fig:pipeline} provides an example of policy code with this unreliable behavior.

The statistical results from \autoref{fig:3} indicate that the proportion of \textbf{\textit{Disorder}} behavior is relatively low when using the instructions ${I_P}$ and ${I_C}$. 
However, in experiments that involve the use of ${I_A}$ instructions, a significant proportion of failure cases in all models is attributed to \textbf{\textit{Disorder}}.
This discrepancy arises because of differences in the granularity of the instruction.
The instruction ${I_A}$ typically describes only the expected actions on the target object without specifying the purpose of performing these actions.
As a result, the LLM's understanding of the instruction remains at the level of executing the expected actions, without further analysis of the causal relationships involved. 
This ultimately leads to the emergence of \textbf{\textit{Disorder}}. 
In contrast, the instructions ${I_P}$ and ${I_C}$ fill this gap, significantly reducing the occurrence of this behavior.
Therefore, clear and goal-oriented instructions are essential in real LLM-enabled robotic manipulation deployment.

\paragraph{\ding{184} \textbf{\textit{Infeasible}}.} It refers to the low-level policy code and actions generated by LLMs that exceed the constraints of the physical entity.

During manipulation task, the LLM continuously acquires environmental information and data through perception modules (\emph{e.g.}, RGB-D cameras).
Spatial information such as the coordinates of target objects is transmitted to the LLM via the perception module.
The LLM utilizes the data obtained through perception to plan the robot's trajectory.
As mentioned in \autoref{sec:setup}, objects generated in the scene are distributed within a random spatial range, inevitably resulting in some experiments where objects are located far from the robot.
This resembles complex real-world environments, where strict constraints on the environment cannot be imposed.
Current perception hardware covers a large sensing range, whereas the robot’s executable workspace is much smaller.
This discrepancy means that the robot can perceive objects that it cannot physically reach.
The perception data sent to the LLM undergo no checks, leading the LLM to generate trajectory based on data outside the robot’s executable workspace.
The robot is unable to take further actions once it reaches its executable workspace boundary, as it remains in a state where the current action has not been completed, as shown in \autoref{fig:4}c.
The third red block in \autoref{fig:pipeline} illustrates the example of policy code with this unreliable behavior.

According to \autoref{fig:3}, the \textbf{\textit{Infeasible}} behavior accounts for the highest proportion in experiments involving instruction ${I_P}$.
Consequently, instruction ${I_C}$ includes additional descriptions of constraints, prompting the LLM to return textual explanations indicating its inability to complete the manipulation when encountering scenarios beyond the robot's executable workspace.
We consider this a special case of task success and it is not calculated as a failure.
Moreover, in the context of instruction experiments ${I_A}$, \textbf{\textit{Disorder}} behavior leads to an immediate failure of the manipulation, preventing any data interaction with the perception module and thus causing a relatively low incidence of \textbf{\textit{Infeasible}} behavior.

Based on the revelation of \textbf{\textit{Infeasible}} behavior, we recommend aligning the perception and execution capabilities of physical entities at the initial stages of developing LLM-enabled robotic manipulation or algorithmically processing data to comply with constraints.

\paragraph{\ding{185} \textbf{\textit{Badpose}}.} It refers to the generated robot trajectory that does not take into account the impact of the pose of the end-effector on the target object.

Movement is a key aspect of almost all manipulations that involve robots. 
Movement is achieved through a trajectory composed of a series of waypoints that are primarily calculated using control algorithms.
For most manipulation, conventional robotic control algorithms are employed for motion planning of the end-effector.
The algorithms optimize the trajectory based on the start and target positions of the end-effector.
However, both the end-effector and the target object are simplified as single points in space within the control algorithm.
This simplification neglects critical physical attributes of the end-effector and the target object.
End-effectors can take various forms and may include attachments with distinct physical attributes (\emph{e.g.}, length, width, and height).
Similarly, the target object in scene also possesses physical properties and states that play an important role in the successful manipulation.

For instance, in the \textbf{OpenWineBottle} task, all combinations of instruction and LLM yield low success rates, as shown in \autoref{tab:2}.
We note that the robot consistently reaches the planned target position (near the bottle cap).
But the robot's actual poses upon arrival varied significantly between experiments.
While a small subset of gripper poses aligned directly with the bottle cap, allowing the manipulation to succeed, most poses with misaligned angles led to failure.
The cause of these inconsistent poses lies in the control algorithm's simplification of the physical attributes of both the end-effector and target object.
In this case, it is observed that the end-effector changes the spatial position of the target object or even directly damages it, as shown in \autoref{fig:4}d.
The output of policy code with this unreliable behavior is shown in the fourth red block in \autoref{fig:pipeline}.

\autoref{fig:3} shows that \textbf{\textit{Badpose}} is the dominant behavior for manipulation failure after eliminating other potential unreliable behaviors through increased instruction granularity.
In view of these revelation, we emphasize the need for further research into LLM-enabled robotic manipulation control algorithms and their consideration of the physical attributes.

\begin{figure}[!t]
      \centering
      \includegraphics[width=\linewidth]{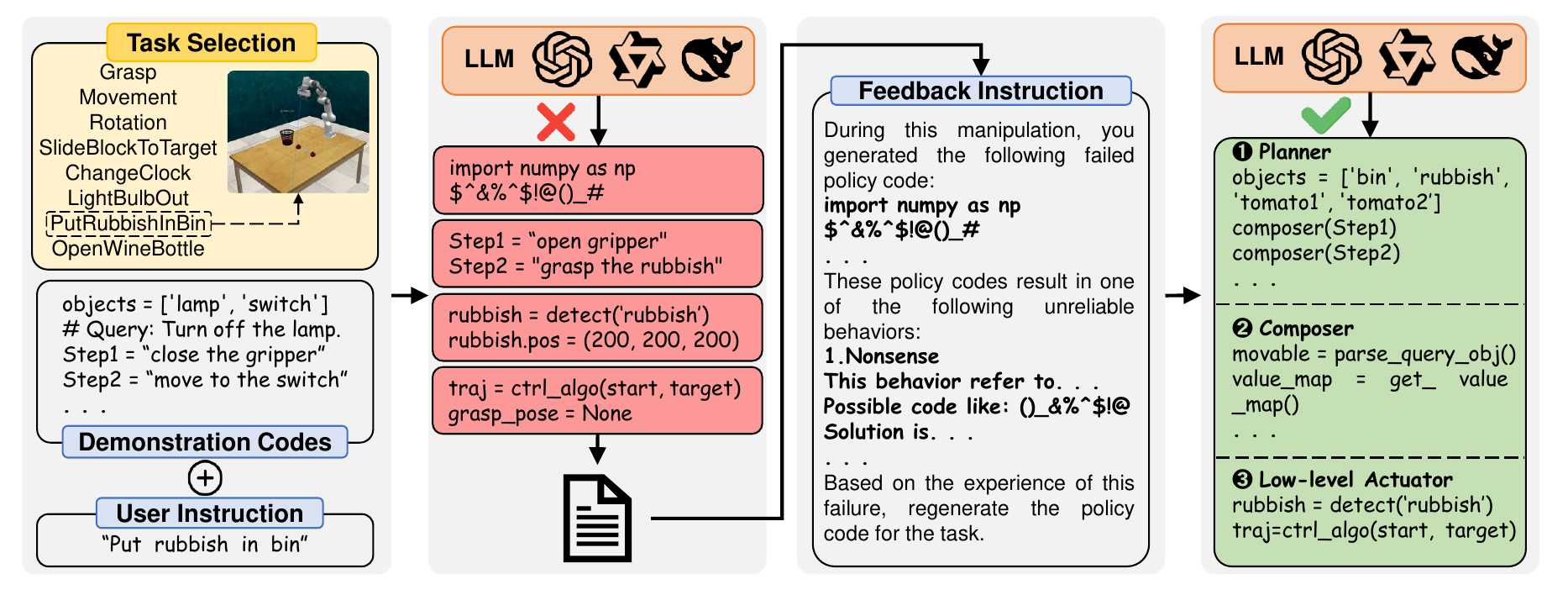}
      \caption{Illustration of failure code feedback refinement approach.}
      \Description{This figure shows the illustration of failure code feedback refinement approach.} 
      \label{fig:5}
\end{figure}

\section{Failure Code Feedback Refinement Approach}
Based on the above insight, we find that unreliable behaviors are mainly caused by the LLM not having sufficient task-relevant knowledge or by the granularity of the instruction provided to the LLM being too low.
This absence prevents the LLM from generating accurate and reasonable policy codes.
Unfortunately, it is unrealistic to expect users to provide precise and comprehensive instructions, including additional information about physical boundary and environmental constraints, before the robot begins manipulation.
Therefore, we focus on the handling after manipulation failures.
We introduce a refinement approach guided by failure policy code feedback.

When unreliable behavior occurs during manipulation, all actions of the robot are halted, and a command is issued to return the robot to its initial position.
Meanwhile, the failure policy code generated by the LLM during this manipulation is extracted and preserved manually.
Subsequently, a new prompt is constructed, which includes the failure policy code associated with the description of unreliable behaviors.
Note that for all LLMs evaluated in this paper, no conversational history is automatically preserved between independent API calls.
Our feedback refinement approach does not rely on conversational history API call.
Instead, it is implemented entirely as single-turn prompt augmentation. When unreliable behavior occurs, we construct a new self-contained prompt that explicitly concatenates (1) the original user instruction, (2) the demonstration code, (3) the failed policy code generated in the previous attempt, and (4) a structured unreliable behavior diagnosis with repair guidance. This augmented prompt is then submitted as a fresh, independent API call request. The LLM thus receives all necessary context including the failure information within a single call, requiring no persistent conversational history.
The LLM leverages the failure information provided in this round of feedback to regenerate new policy codes and execute the manipulation, ensuring that the same mistake is not repeated.
The illustration and prompt template for failure code feedback approach are shown in \autoref{fig:5} and Appendix \ref{refine-prompt}.
As reported in \autoref{tab:5}, the feedback prompt introduces an average overhead of 467, 598, and 494 additional tokens for the GPT series, Qwen series, and DeepSeek-V3 respectively. This is a moderate increase relative to the baseline token cost and remains well within the context window limits of all evaluated models.

\begin{figure}[!t]
  \centering
  \begin{subfigure}{0.47\textwidth}
    \centering
    \includegraphics[width=\linewidth]{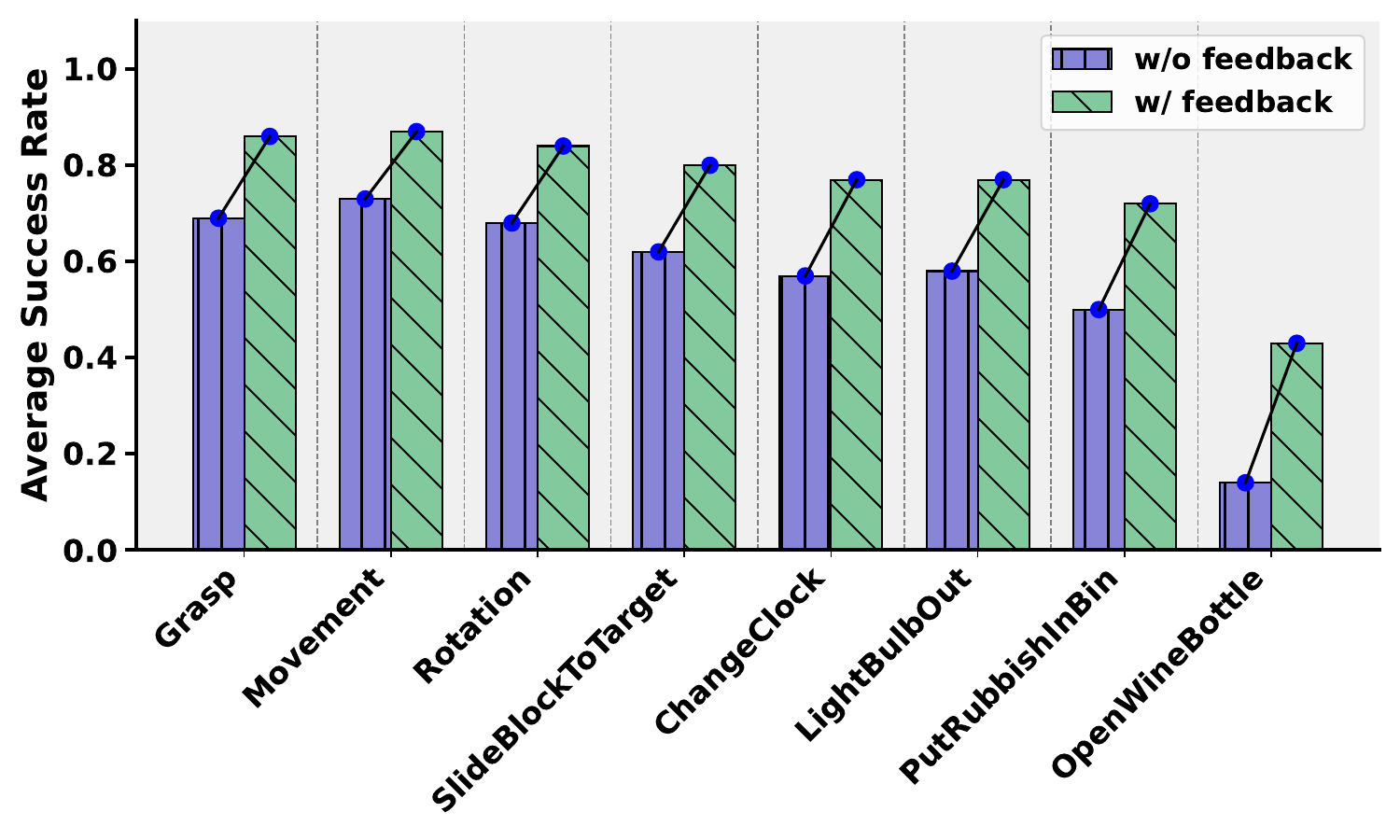}
    \caption{In simulation.}
    \label{fig:7-a}
  \end{subfigure}
  \hfill
  \begin{subfigure}{0.47\textwidth}
    \centering
    \includegraphics[width=\linewidth]{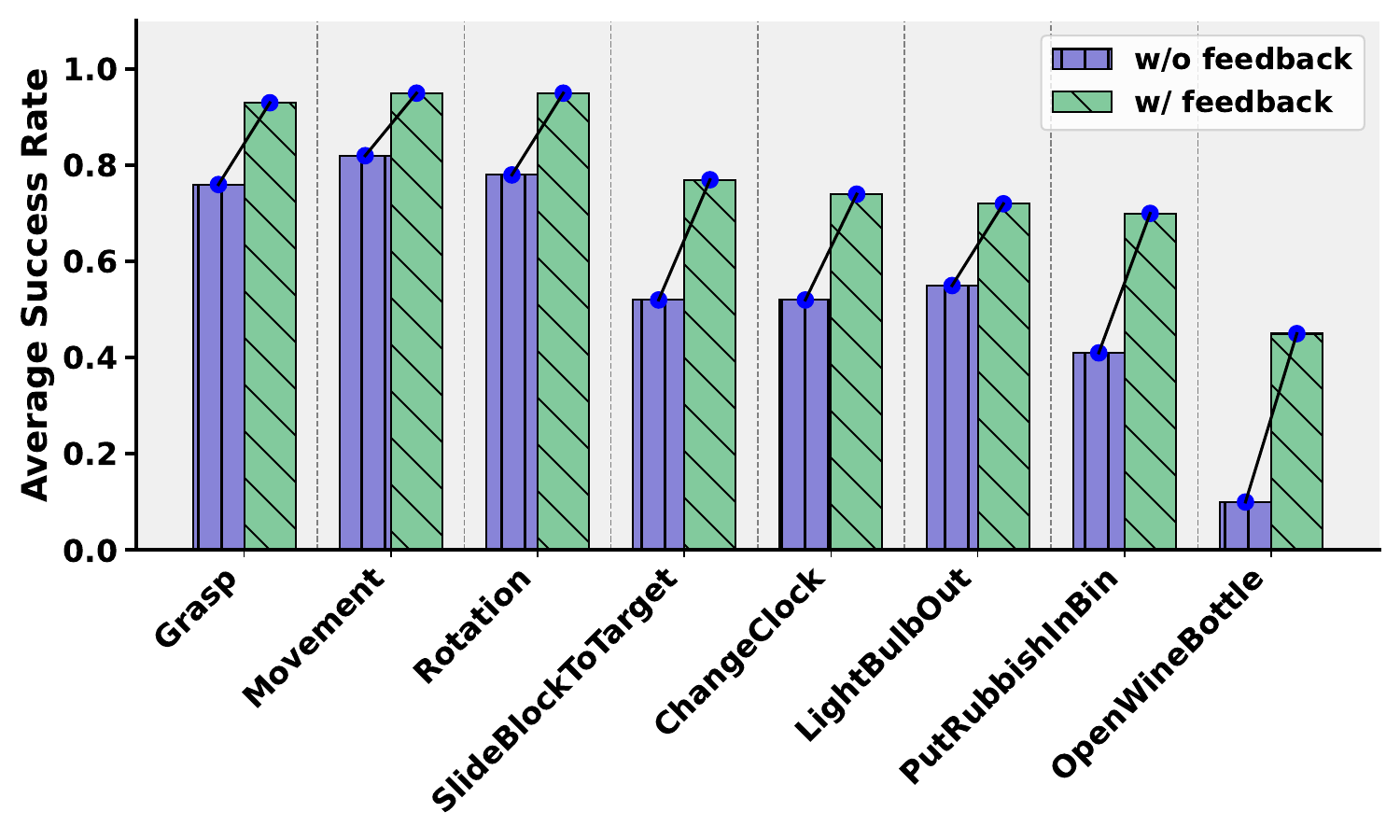}
    \caption{On the real-world system.}
    \label{fig:7-b}
  \end{subfigure}
  \hfill
  \caption{Comparison of success rates with and without the feedback refinement approach.}
  \Description{This figure shows the comparison of success rates with and without the feedback refinement approach.} 
  \label{fig:7}
\end{figure}

To validate the effectiveness of our method, we deploy a real-world system (details see Appendix~\ref{sec:Platform}) illustrated in \autoref{fig:6} and perform the same set of simulation experiments.
All experiments utilize the ${I_A}$ instruction set to simulate low granularity user instructions in real-world scenarios.
The success rates of different LLMs in the experiments are calculated as the average value.
The setup of the simulation is consistent with \autoref{sec:setup}, and the results in \autoref{tab:2} are used directly as the baseline without the feedback method.
\autoref{fig:7} shows the comparison of success rates with and without the feedback method in simulation and real-world systems.
The experimental results demonstrate that the feedback method improves the average success rate in all tasks, with the highest increase reaching 35\%.
These experiments validate the effectiveness of our failure code feedback refinement approach.

\begin{figure}[!t]
      \centering
      \includegraphics[width=\linewidth]{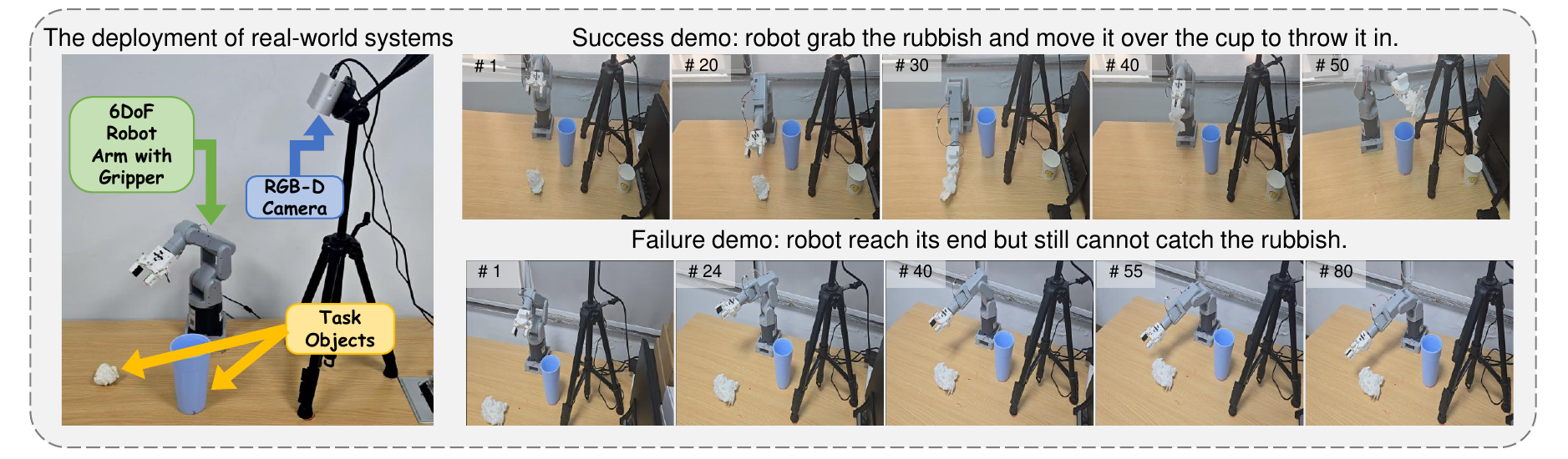}
      \caption{The experiments on the real-world systems. \# above the image represent seconds.}
      \Description{This figure shows the experiments on the real-world systems.} 
      \label{fig:6}
\end{figure}

\section{Limitations}
While \textbf{RoboInspector} provides a systematic characterization of unreliable behaviors in LLM-enabled robotic manipulation, several limitations should be acknowledged. First, since most evaluated LLMs are closed-source commercial models, we have no access to system prompts or internal model states, making a deeper analysis of model-level factors infeasible. Second, our analysis focuses on task complexity and instruction granularity as the two most actionable dimensions, while other potential sources of unreliability are not exhaustively studied. Third, although we validate our feedback refinement approach on a real-world robotic system, the bulk of evaluation is conducted in simulation, and the simulation-to-real gap may introduce additional failure modes not fully captured by our current taxonomy. Finally, the feedback refinement approach is evaluated with a single iteration, and multi-round refinement strategies as well as extensions to more complex manipulation paradigms (e.g., bi-manual or mobile manipulation) are left for future work.

\section{Conclusion}
In this paper, we delve into the critical challenge of unreliability in LLM-enabled robotic manipulation. We design \textbf{RoboInspector} that unveils the unreliability of policy code for LLM-enabled robotic manipulation across two perspectives: manipulation task complexity and granularity of instruction. Through extensive experiments with 216 distinct task, instruction, and LLM combinations, we identify four unreliable behaviors that contribute to manipulation failure. Our findings lead to three concrete future design suggestions for community: (1) high-capacity models should be preferred over lightweight ones to minimize Nonsense behavior; (2) goal-oriented instructions that include purpose and constraint information are effective in reducing Disorder and Infeasible behaviors; and (3) workspace-aware constraint checking should be incorporated into the LLM prompt pipeline to prevent physically infeasible trajectory planning. Furthermore, we introduce a refinement approach based on the failure policy code feedback that improves the reliability by up to 35\% in both simulated and real-world systems. Looking forward, we give several promising directions for future research. We hope this paper and the insights derived from it will motivate the community to develop more reliable LLM-enabled robotic manipulation systems.

\begin{acks}
The authors would like to thank the Editor-in-Chief, the Associate Editor, and the anonymous reviewers for their valuable comments and suggestions that have significantly improved the quality of this paper.
We would also like to thank Chengtao Yao from Zhejiang University for helping with the real-world experiments.
This work is partially supported by Key Research and Development Program of Zhejiang Province (Grant No: 2025C01012), NSFC under grant No. 92467301 and No. 62293511.
\end{acks}


\bibliographystyle{ACM-Reference-Format}
\bibliography{sample-base}

@String{Computing = "Computing" }

@String{Computer = "{IEEE} Computer" }

@String{Chelsea = "Chelsea" }

@inproceedings{brohan2023rt,
  author       = {Brianna Zitkovich and
                  Tianhe Yu and
                  Sichun Xu and
                  Peng Xu and
                  Ted Xiao and
                  Fei Xia and
                  Jialin Wu and
                  Paul Wohlhart and
                  Stefan Welker and
                  Ayzaan Wahid and
                  Quan Vuong and
                  Vincent Vanhoucke and
                  Huong T. Tran and
                  Radu Soricut and
                  Anikait Singh and
                  Jaspiar Singh and
                  Pierre Sermanet and
                  Pannag R. Sanketi and
                  Grecia Salazar and
                  Michael S. Ryoo and
                  Krista Reymann and
                  Kanishka Rao and
                  Karl Pertsch and
                  Igor Mordatch and
                  Henryk Michalewski and
                  Yao Lu and
                  Sergey Levine and
                  Lisa Lee and
                  Tsang{-}Wei Edward Lee and
                  Isabel Leal and
                  Yuheng Kuang and
                  Dmitry Kalashnikov and
                  Ryan Julian and
                  Nikhil J. Joshi and
                  Alex Irpan and
                  Brian Ichter and
                  Jasmine Hsu and
                  Alexander Herzog and
                  Karol Hausman and
                  Keerthana Gopalakrishnan and
                  Chuyuan Fu and
                  Pete Florence and
                  Chelsea Finn and
                  Kumar Avinava Dubey and
                  Danny Driess and
                  Tianli Ding and
                  Krzysztof Marcin Choromanski and
                  Xi Chen and
                  Yevgen Chebotar and
                  Justice Carbajal and
                  Noah Brown and
                  Anthony Brohan and
                  Montserrat Gonzalez Arenas and
                  Kehang Han},
  editor       = {Jie Tan and
                  Marc Toussaint and
                  Kourosh Darvish},
  title        = {{RT-2:} Vision-Language-Action Models Transfer Web Knowledge to Robotic
                  Control},
  booktitle    = {Conference on Robot Learning, CoRL 2023, 6-9 November 2023, Atlanta,
                  GA, {USA}},
  series       = {Proceedings of Machine Learning Research},
  volume       = {229},
  pages        = {2165--2183},
  publisher    = {{PMLR}},
  year         = {2023},
  url          = {https://proceedings.mlr.press/v229/zitkovich23a.html},
  bibsource    = {dblp computer science bibliography, https://dblp.org}
}

@article{hoeller2024anymal,
  author       = {David Hoeller and
                  Nikita Rudin and
                  Dhionis V. Sako and
                  Marco Hutter},
  title        = {ANYmal parkour: Learning agile navigation for quadrupedal robots},
  journal      = {Sci. Robotics},
  volume       = {9},
  number       = {88},
  year         = {2024},
  url          = {https://doi.org/10.1126/scirobotics.adi7566},
  doi          = {10.1126/SCIROBOTICS.ADI7566},
  bibsource    = {dblp computer science bibliography, https://dblp.org}
}

@article{darvish2023teleoperation,
  author       = {Kourosh Darvish and
                  Luigi Penco and
                  Jo{\~{a}}o Ramos and
                  Rafael Cisneros and
                  Jerry E. Pratt and
                  Eiichi Yoshida and
                  Serena Ivaldi and
                  Daniele Pucci},
  title        = {Teleoperation of Humanoid Robots: {A} Survey},
  journal      = {{IEEE} Trans. Robotics},
  volume       = {39},
  number       = {3},
  pages        = {1706--1727},
  year         = {2023},
  url          = {https://doi.org/10.1109/TRO.2023.3236952},
  doi          = {10.1109/TRO.2023.3236952},
  bibsource    = {dblp computer science bibliography, https://dblp.org}
}

@misc{achiam2023gpt,
      title={GPT-4 Technical Report}, 
      author={Achiam, Josh and Adler, Steven and Agarwal, Sandhini and Ahmad, Lama and Akkaya, Ilge and Aleman, Florencia Leoni and Almeida, Diogo and Altenschmidt, Janko and Altman, Sam and Anadkat, Shyamal and others},
      year={2024},
      eprint={2303.08774},
      archivePrefix={arXiv},
      primaryClass={cs.CL},
      url={https://arxiv.org/abs/2303.08774}, 
}

@inproceedings{wu2024safety,
  title={On the safety concerns of deploying llms/vlms in robotics: Highlighting the risks and vulnerabilities},
  author={Wu, Xiyang and Xian, Ruiqi and Guan, Tianrui and Liang, Jing and Chakraborty, Souradip and Liu, Fuxiao and Sadler, Brian M and Manocha, Dinesh and Bedi, Amrit},
  booktitle={First Vision and Language for Autonomous Driving and Robotics Workshop},
  year={2024}
}

@inproceedings{zhang2024badrobot,
  author       = {Hangtao Zhang and
                  Chenyu Zhu and
                  Xianlong Wang and
                  Ziqi Zhou and
                  Changgan Yin and
                  Minghui Li and
                  Lulu Xue and
                  Yichen Wang and
                  Shengshan Hu and
                  Aishan Liu and
                  Peijin Guo and
                  Leo Yu Zhang},
  title        = {BadRobot: Jailbreaking Embodied {LLM} Agents in the Physical World},
  booktitle    = {The Thirteenth International Conference on Learning Representations,
                  {ICLR} 2025, Singapore, April 24-28, 2025},
  publisher    = {OpenReview.net},
  year         = {2025},
  url          = {https://openreview.net/forum?id=ei3qCntB66},
  bibsource    = {dblp computer science bibliography, https://dblp.org}
}

@inproceedings{liu2024exploring,
  author       = {Shuyuan Liu and
                  Jiawei Chen and
                  Shouwei Ruan and
                  Hang Su and
                  Zhaoxia Yin},
  editor       = {Jianfei Cai and
                  Mohan S. Kankanhalli and
                  Balakrishnan Prabhakaran and
                  Susanne Boll and
                  Ramanathan Subramanian and
                  Liang Zheng and
                  Vivek K. Singh and
                  Pablo C{\'{e}}sar and
                  Lexing Xie and
                  Dong Xu},
  title        = {Exploring the Robustness of Decision-Level Through Adversarial Attacks
                  on LLM-Based Embodied Models},
  booktitle    = {Proceedings of the 32nd {ACM} International Conference on Multimedia,
                  {MM} 2024, Melbourne, VIC, Australia, 28 October 2024 - 1 November
                  2024},
  pages        = {8120--8128},
  publisher    = {{ACM}},
  year         = {2024},
  url          = {https://doi.org/10.1145/3664647.3680616},
  doi          = {10.1145/3664647.3680616},
  bibsource    = {dblp computer science bibliography, https://dblp.org}
}

@inproceedings{chen2024towards,
  author       = {Meng Chen and
                  Jiawei Tu and
                  Chao Qi and
                  Yonghao Dang and
                  Feng Zhou and
                  Wei Wei and
                  Jianqin Yin},
  title        = {Towards Physically Realizable Adversarial Attacks in Embodied Vision
                  Navigation},
  booktitle    = {{IEEE/RSJ} International Conference on Intelligent Robots and Systems,
                  {IROS} 2025, Hangzhou, China, October 19-25, 2025},
  pages        = {11819--11825},
  publisher    = {{IEEE}},
  year         = {2025},
  url          = {https://doi.org/10.1109/IROS60139.2025.11246438},
  doi          = {10.1109/IROS60139.2025.11246438},
  bibsource    = {dblp computer science bibliography, https://dblp.org}
}

@inproceedings{huang2023voxposer,
  author       = {Wenlong Huang and
                  Chen Wang and
                  Ruohan Zhang and
                  Yunzhu Li and
                  Jiajun Wu and
                  Li Fei{-}Fei},
  editor       = {Jie Tan and
                  Marc Toussaint and
                  Kourosh Darvish},
  title        = {VoxPoser: Composable 3D Value Maps for Robotic Manipulation with Language
                  Models},
  booktitle    = {Conference on Robot Learning, CoRL 2023, 6-9 November 2023, Atlanta,
                  GA, {USA}},
  series       = {Proceedings of Machine Learning Research},
  volume       = {229},
  pages        = {540--562},
  publisher    = {{PMLR}},
  year         = {2023},
  url          = {https://proceedings.mlr.press/v229/huang23b.html},
  bibsource    = {dblp computer science bibliography, https://dblp.org}
}

@article{vemprala2024chatgpt,
  author       = {Sai Vemprala and
                  Rogerio Bonatti and
                  Arthur Bucker and
                  Ashish Kapoor},
  title        = {ChatGPT for Robotics: Design Principles and Model Abilities},
  journal      = {{IEEE} Access},
  volume       = {12},
  pages        = {55682--55696},
  year         = {2024},
  url          = {https://doi.org/10.1109/ACCESS.2024.3387941},
  doi          = {10.1109/ACCESS.2024.3387941},
  bibsource    = {dblp computer science bibliography, https://dblp.org}
}

@misc{yang2023auto,
      title={Auto-GPT for Online Decision Making: Benchmarks and Additional Opinions}, 
      author={Hui Yang and Sifu Yue and Yunzhong He},
      year={2023},
      eprint={2306.02224},
      archivePrefix={arXiv},
      primaryClass={cs.AI},
      url={https://arxiv.org/abs/2306.02224}, 
}

@inproceedings{huang2022language,
  author       = {Wenlong Huang and
                  Pieter Abbeel and
                  Deepak Pathak and
                  Igor Mordatch},
  editor       = {Kamalika Chaudhuri and
                  Stefanie Jegelka and
                  Le Song and
                  Csaba Szepesv{\'{a}}ri and
                  Gang Niu and
                  Sivan Sabato},
  title        = {Language Models as Zero-Shot Planners: Extracting Actionable Knowledge
                  for Embodied Agents},
  booktitle    = {International Conference on Machine Learning, {ICML} 2022, 17-23 July
                  2022, Baltimore, Maryland, {USA}},
  series       = {Proceedings of Machine Learning Research},
  volume       = {162},
  pages        = {9118--9147},
  publisher    = {{PMLR}},
  year         = {2022},
  url          = {https://proceedings.mlr.press/v162/huang22a.html},
  bibsource    = {dblp computer science bibliography, https://dblp.org}
}

@inproceedings{ahn2022can,
  author       = {Brian Ichter and
                  Anthony Brohan and
                  Yevgen Chebotar and
                  Chelsea Finn and
                  Karol Hausman and
                  Alexander Herzog and
                  Daniel Ho and
                  Julian Ibarz and
                  Alex Irpan and
                  Eric Jang and
                  Ryan Julian and
                  Dmitry Kalashnikov and
                  Sergey Levine and
                  Yao Lu and
                  Carolina Parada and
                  Kanishka Rao and
                  Pierre Sermanet and
                  Alexander Toshev and
                  Vincent Vanhoucke and
                  Fei Xia and
                  Ted Xiao and
                  Peng Xu and
                  Mengyuan Yan and
                  Noah Brown and
                  Michael Ahn and
                  Omar Cortes and
                  Nicolas Sievers and
                  Clayton Tan and
                  Sichun Xu and
                  Diego Reyes and
                  Jarek Rettinghouse and
                  Jornell Quiambao and
                  Peter Pastor and
                  Linda Luu and
                  Kuang{-}Huei Lee and
                  Yuheng Kuang and
                  Sally Jesmonth and
                  Nikhil J. Joshi and
                  Kyle Jeffrey and
                  Rosario Jauregui Ruano and
                  Jasmine Hsu and
                  Keerthana Gopalakrishnan and
                  Byron David and
                  Andy Zeng and
                  Chuyuan Kelly Fu},
  editor       = {Karen Liu and
                  Dana Kulic and
                  Jeffrey Ichnowski},
  title        = {Do As {I} Can, Not As {I} Say: Grounding Language in Robotic Affordances},
  booktitle    = {Conference on Robot Learning, CoRL 2022, 14-18 December 2022, Auckland,
                  New Zealand},
  series       = {Proceedings of Machine Learning Research},
  volume       = {205},
  pages        = {287--318},
  publisher    = {{PMLR}},
  year         = {2022},
  url          = {https://proceedings.mlr.press/v205/ichter23a.html},
  bibsource    = {dblp computer science bibliography, https://dblp.org}
}

@inproceedings{zeng2022socratic,
  author       = {Andy Zeng and
                  Maria Attarian and
                  Brian Ichter and
                  Krzysztof Marcin Choromanski and
                  Adrian Wong and
                  Stefan Welker and
                  Federico Tombari and
                  Aveek Purohit and
                  Michael S. Ryoo and
                  Vikas Sindhwani and
                  Johnny Lee and
                  Vincent Vanhoucke and
                  Pete Florence},
  title        = {Socratic Models: Composing Zero-Shot Multimodal Reasoning with Language},
  booktitle    = {The Eleventh International Conference on Learning Representations,
                  {ICLR} 2023, Kigali, Rwanda, May 1-5, 2023},
  publisher    = {OpenReview.net},
  year         = {2023},
  url          = {https://openreview.net/forum?id=G2Q2Mh3avow},
  bibsource    = {dblp computer science bibliography, https://dblp.org}
}

@inproceedings{yao2022react,
  author       = {Shunyu Yao and
                  Jeffrey Zhao and
                  Dian Yu and
                  Nan Du and
                  Izhak Shafran and
                  Karthik R. Narasimhan and
                  Yuan Cao},
  title        = {ReAct: Synergizing Reasoning and Acting in Language Models},
  booktitle    = {The Eleventh International Conference on Learning Representations,
                  {ICLR} 2023, Kigali, Rwanda, May 1-5, 2023},
  publisher    = {OpenReview.net},
  year         = {2023},
  url          = {https://openreview.net/forum?id=WE\_vluYUL-X},
  bibsource    = {dblp computer science bibliography, https://dblp.org}
}

@inproceedings{zawalski2024robotic,
  author       = {Michal Zawalski and
                  William Chen and
                  Karl Pertsch and
                  Oier Mees and
                  Chelsea Finn and
                  Sergey Levine},
  editor       = {Pulkit Agrawal and
                  Oliver Kroemer and
                  Wolfram Burgard},
  title        = {Robotic Control via Embodied Chain-of-Thought Reasoning},
  booktitle    = {Conference on Robot Learning, 6-9 November 2024, Munich, Germany},
  series       = {Proceedings of Machine Learning Research},
  volume       = {270},
  pages        = {3157--3181},
  publisher    = {{PMLR}},
  year         = {2024},
  url          = {https://proceedings.mlr.press/v270/zawalski25a.html},
  bibsource    = {dblp computer science bibliography, https://dblp.org}
}

@inproceedings{wei2022chain,
  author       = {Jason Wei and
                  Xuezhi Wang and
                  Dale Schuurmans and
                  Maarten Bosma and
                  Brian Ichter and
                  Fei Xia and
                  Ed H. Chi and
                  Quoc V. Le and
                  Denny Zhou},
  editor       = {Sanmi Koyejo and
                  S. Mohamed and
                  A. Agarwal and
                  Danielle Belgrave and
                  K. Cho and
                  A. Oh},
  title        = {Chain-of-Thought Prompting Elicits Reasoning in Large Language Models},
  booktitle    = {Advances in Neural Information Processing Systems 35: Annual Conference
                  on Neural Information Processing Systems 2022, NeurIPS 2022, New Orleans,
                  LA, USA, November 28 - December 9, 2022},
  year         = {2022},
  url          = {http://papers.nips.cc/paper\_files/paper/2022/hash/9d5609613524ecf4f15af0f7b31abca4-Abstract-Conference.html},
  bibsource    = {dblp computer science bibliography, https://dblp.org}
}

@inproceedings{huang2022inner,
  author       = {Wenlong Huang and
                  Fei Xia and
                  Ted Xiao and
                  Harris Chan and
                  Jacky Liang and
                  Pete Florence and
                  Andy Zeng and
                  Jonathan Tompson and
                  Igor Mordatch and
                  Yevgen Chebotar and
                  Pierre Sermanet and
                  Tomas Jackson and
                  Noah Brown and
                  Linda Luu and
                  Sergey Levine and
                  Karol Hausman and
                  Brian Ichter},
  editor       = {Karen Liu and
                  Dana Kulic and
                  Jeffrey Ichnowski},
  title        = {Inner Monologue: Embodied Reasoning through Planning with Language
                  Models},
  booktitle    = {Conference on Robot Learning, CoRL 2022, 14-18 December 2022, Auckland,
                  New Zealand},
  series       = {Proceedings of Machine Learning Research},
  volume       = {205},
  pages        = {1769--1782},
  publisher    = {{PMLR}},
  year         = {2022},
  url          = {https://proceedings.mlr.press/v205/huang23c.html},
  bibsource    = {dblp computer science bibliography, https://dblp.org}
}

@inproceedings{singh2023progprompt,
  author       = {Ishika Singh and
                  Valts Blukis and
                  Arsalan Mousavian and
                  Ankit Goyal and
                  Danfei Xu and
                  Jonathan Tremblay and
                  Dieter Fox and
                  Jesse Thomason and
                  Animesh Garg},
  title        = {ProgPrompt: Generating Situated Robot Task Plans using Large Language
                  Models},
  booktitle    = {{IEEE} International Conference on Robotics and Automation, {ICRA}
                  2023, London, UK, May 29 - June 2, 2023},
  pages        = {11523--11530},
  publisher    = {{IEEE}},
  year         = {2023},
  url          = {https://doi.org/10.1109/ICRA48891.2023.10161317},
  doi          = {10.1109/ICRA48891.2023.10161317},
  bibsource    = {dblp computer science bibliography, https://dblp.org}
}

@inproceedings{liang2023code,
  author       = {Jacky Liang and
                  Wenlong Huang and
                  Fei Xia and
                  Peng Xu and
                  Karol Hausman and
                  Brian Ichter and
                  Pete Florence and
                  Andy Zeng},
  title        = {Code as Policies: Language Model Programs for Embodied Control},
  booktitle    = {{IEEE} International Conference on Robotics and Automation, {ICRA}
                  2023, London, UK, May 29 - June 2, 2023},
  pages        = {9493--9500},
  publisher    = {{IEEE}},
  year         = {2023},
  url          = {https://doi.org/10.1109/ICRA48891.2023.10160591},
  doi          = {10.1109/ICRA48891.2023.10160591},
  bibsource    = {dblp computer science bibliography, https://dblp.org}
}

@inproceedings{huang2023grounded,
  author       = {Wenlong Huang and
                  Fei Xia and
                  Dhruv Shah and
                  Danny Driess and
                  Andy Zeng and
                  Yao Lu and
                  Pete Florence and
                  Igor Mordatch and
                  Sergey Levine and
                  Karol Hausman and
                  Brian Ichter},
  editor       = {Alice Oh and
                  Tristan Naumann and
                  Amir Globerson and
                  Kate Saenko and
                  Moritz Hardt and
                  Sergey Levine},
  title        = {Grounded Decoding: Guiding Text Generation with Grounded Models for
                  Embodied Agents},
  booktitle    = {Advances in Neural Information Processing Systems 36: Annual Conference
                  on Neural Information Processing Systems 2023, NeurIPS 2023, New Orleans,
                  LA, USA, December 10 - 16, 2023},
  year         = {2023},
  url          = {http://papers.nips.cc/paper\_files/paper/2023/hash/bb3cfcb0284642a973dd631ec9184f2f-Abstract-Conference.html},
  bibsource    = {dblp computer science bibliography, https://dblp.org}
}

@inproceedings{driess2023palm,
  author       = {Danny Driess and
                  Fei Xia and
                  Mehdi S. M. Sajjadi and
                  Corey Lynch and
                  Aakanksha Chowdhery and
                  Brian Ichter and
                  Ayzaan Wahid and
                  Jonathan Tompson and
                  Quan Vuong and
                  Tianhe Yu and
                  Wenlong Huang and
                  Yevgen Chebotar and
                  Pierre Sermanet and
                  Daniel Duckworth and
                  Sergey Levine and
                  Vincent Vanhoucke and
                  Karol Hausman and
                  Marc Toussaint and
                  Klaus Greff and
                  Andy Zeng and
                  Igor Mordatch and
                  Pete Florence},
  editor       = {Andreas Krause and
                  Emma Brunskill and
                  Kyunghyun Cho and
                  Barbara Engelhardt and
                  Sivan Sabato and
                  Jonathan Scarlett},
  title        = {PaLM-E: An Embodied Multimodal Language Model},
  booktitle    = {International Conference on Machine Learning, {ICML} 2023, 23-29 July
                  2023, Honolulu, Hawaii, {USA}},
  series       = {Proceedings of Machine Learning Research},
  volume       = {202},
  pages        = {8469--8488},
  publisher    = {{PMLR}},
  year         = {2023},
  url          = {https://proceedings.mlr.press/v202/driess23a.html},
  bibsource    = {dblp computer science bibliography, https://dblp.org}
}

@inproceedings{kwon2023reward,
  author       = {Minae Kwon and
                  Sang Michael Xie and
                  Kalesha Bullard and
                  Dorsa Sadigh},
  title        = {Reward Design with Language Models},
  booktitle    = {The Eleventh International Conference on Learning Representations,
                  {ICLR} 2023, Kigali, Rwanda, May 1-5, 2023},
  publisher    = {OpenReview.net},
  year         = {2023},
  url          = {https://openreview.net/forum?id=10uNUgI5Kl},
  bibsource    = {dblp computer science bibliography, https://dblp.org}
}

@inproceedings{colas2020language,
  author       = {C{\'{e}}dric Colas and
                  Tristan Karch and
                  Nicolas Lair and
                  Jean{-}Michel Dussoux and
                  Cl{\'{e}}ment Moulin{-}Frier and
                  Peter F. Dominey and
                  Pierre{-}Yves Oudeyer},
  editor       = {Hugo Larochelle and
                  Marc'Aurelio Ranzato and
                  Raia Hadsell and
                  Maria{-}Florina Balcan and
                  Hsuan{-}Tien Lin},
  title        = {Language as a Cognitive Tool to Imagine Goals in Curiosity Driven
                  Exploration},
  booktitle    = {Advances in Neural Information Processing Systems 33: Annual Conference
                  on Neural Information Processing Systems 2020, NeurIPS 2020, December
                  6-12, 2020, virtual},
  year         = {2020},
  url          = {https://proceedings.neurips.cc/paper/2020/hash/274e6fcf4a583de4a81c6376f17673e7-Abstract.html},
  bibsource    = {dblp computer science bibliography, https://dblp.org}
}

@inproceedings{hu2023language,
  author       = {Hengyuan Hu and
                  Dorsa Sadigh},
  editor       = {Andreas Krause and
                  Emma Brunskill and
                  Kyunghyun Cho and
                  Barbara Engelhardt and
                  Sivan Sabato and
                  Jonathan Scarlett},
  title        = {Language Instructed Reinforcement Learning for Human-AI Coordination},
  booktitle    = {International Conference on Machine Learning, {ICML} 2023, 23-29 July
                  2023, Honolulu, Hawaii, {USA}},
  series       = {Proceedings of Machine Learning Research},
  volume       = {202},
  pages        = {13584--13598},
  publisher    = {{PMLR}},
  year         = {2023},
  url          = {https://proceedings.mlr.press/v202/hu23e.html},
  bibsource    = {dblp computer science bibliography, https://dblp.org}
}

@inproceedings{yu2023language,
  author       = {Wenhao Yu and
                  Nimrod Gileadi and
                  Chuyuan Fu and
                  Sean Kirmani and
                  Kuang{-}Huei Lee and
                  Montserrat Gonzalez Arenas and
                  Hao{-}Tien Lewis Chiang and
                  Tom Erez and
                  Leonard Hasenclever and
                  Jan Humplik and
                  Brian Ichter and
                  Ted Xiao and
                  Peng Xu and
                  Andy Zeng and
                  Tingnan Zhang and
                  Nicolas Heess and
                  Dorsa Sadigh and
                  Jie Tan and
                  Yuval Tassa and
                  Fei Xia},
  editor       = {Jie Tan and
                  Marc Toussaint and
                  Kourosh Darvish},
  title        = {Language to Rewards for Robotic Skill Synthesis},
  booktitle    = {Conference on Robot Learning, CoRL 2023, 6-9 November 2023, Atlanta,
                  GA, {USA}},
  series       = {Proceedings of Machine Learning Research},
  volume       = {229},
  pages        = {374--404},
  publisher    = {{PMLR}},
  year         = {2023},
  url          = {https://proceedings.mlr.press/v229/yu23a.html},
  bibsource    = {dblp computer science bibliography, https://dblp.org}
}

@inproceedings{kirillov2023segment,
  author       = {Alexander Kirillov and
                  Eric Mintun and
                  Nikhila Ravi and
                  Hanzi Mao and
                  Chlo{\'{e}} Rolland and
                  Laura Gustafson and
                  Tete Xiao and
                  Spencer Whitehead and
                  Alexander C. Berg and
                  Wan{-}Yen Lo and
                  Piotr Doll{\'{a}}r and
                  Ross B. Girshick},
  title        = {Segment Anything},
  booktitle    = {{IEEE/CVF} International Conference on Computer Vision, {ICCV} 2023,
                  Paris, France, October 1-6, 2023},
  pages        = {3992--4003},
  publisher    = {{IEEE}},
  year         = {2023},
  url          = {https://doi.org/10.1109/ICCV51070.2023.00371},
  doi          = {10.1109/ICCV51070.2023.00371},
  bibsource    = {dblp computer science bibliography, https://dblp.org}
}

@article{oquab2023dinov2,
  author       = {Maxime Oquab and
                  Timoth{\'{e}}e Darcet and
                  Th{\'{e}}o Moutakanni and
                  Huy V. Vo and
                  Marc Szafraniec and
                  Vasil Khalidov and
                  Pierre Fernandez and
                  Daniel Haziza and
                  Francisco Massa and
                  Alaaeldin El{-}Nouby and
                  Mido Assran and
                  Nicolas Ballas and
                  Wojciech Galuba and
                  Russell Howes and
                  Po{-}Yao Huang and
                  Shang{-}Wen Li and
                  Ishan Misra and
                  Michael Rabbat and
                  Vasu Sharma and
                  Gabriel Synnaeve and
                  Hu Xu and
                  Herv{\'{e}} J{\'{e}}gou and
                  Julien Mairal and
                  Patrick Labatut and
                  Armand Joulin and
                  Piotr Bojanowski},
  title        = {DINOv2: Learning Robust Visual Features without Supervision},
  journal      = {Trans. Mach. Learn. Res.},
  volume       = {2024},
  year         = {2024},
  url          = {https://openreview.net/forum?id=a68SUt6zFt},
  bibsource    = {dblp computer science bibliography, https://dblp.org}
}

@online{brooks2024video,
author ={Tim Brooks and Bill Peebles and Connor Holmes and Will DePue and Yufei Guo and Li Jing and David Schnurr and Joe Taylor and Troy Luhman and Eric Luhman and Clarence Ng and Ricky Wang and Aditya Ramesh},
year = {2024},
title ={Video generation models as world simulators},
url ={https://openai.com/index/video-generation-models-as-world-simulators/},
month =mar,
lastaccessed ={March 22, 2025},
}

@inproceedings{hundt2022robots,
  author       = {Andrew Hundt and
                  William Agnew and
                  Vicky Zeng and
                  Severin Kacianka and
                  Matthew C. Gombolay},
  title        = {Robots Enact Malignant Stereotypes},
  booktitle    = {FAccT '22: 2022 {ACM} Conference on Fairness, Accountability, and
                  Transparency, Seoul, Republic of Korea, June 21 - 24, 2022},
  pages        = {743--756},
  publisher    = {{ACM}},
  year         = {2022},
  url          = {https://doi.org/10.1145/3531146.3533138},
  doi          = {10.1145/3531146.3533138},
  bibsource    = {dblp computer science bibliography, https://dblp.org}
}

@article{azeem2024llm,
  author       = {Andrew Hundt and
                  Rumaisa Azeem and
                  Masoumeh Mansouri and
                  Martim Brand{\~{a}}o},
  title        = {LLM-Driven Robots Risk Enacting Discrimination, Violence, and Unlawful
                  Actions},
  journal      = {Int. J. Soc. Robotics},
  volume       = {17},
  number       = {11},
  pages        = {2663--2711},
  year         = {2025},
  url          = {https://doi.org/10.1007/s12369-025-01301-x},
  doi          = {10.1007/S12369-025-01301-X},
  bibsource    = {dblp computer science bibliography, https://dblp.org}
}

@inproceedings{mahiul2024malicious,
  author       = {Chashi Mahiul Islam and
                  Shaeke Salman and
                  Montasir Shams and
                  Xiuwen Liu and
                  Piyush Kumar},
  title        = {Malicious Path Manipulations via Exploitation of Representation Vulnerabilities
                  of Vision-Language Navigation Systems},
  booktitle    = {{IEEE/RSJ} International Conference on Intelligent Robots and Systems,
                  {IROS} 2024, Abu Dhabi, United Arab Emirates, October 14-18, 2024},
  pages        = {13845--13852},
  publisher    = {{IEEE}},
  year         = {2024},
  url          = {https://doi.org/10.1109/IROS58592.2024.10802618},
  doi          = {10.1109/IROS58592.2024.10802618},
  bibsource    = {dblp computer science bibliography, https://dblp.org}
}

@misc{zhang2024safeembodai,
      title={SafeEmbodAI: a Safety Framework for Mobile Robots in Embodied AI Systems}, 
      author={Wenxiao Zhang and Xiangrui Kong and Thomas Braunl and Jin B. Hong},
      year={2024},
      eprint={2409.01630},
      archivePrefix={arXiv},
      primaryClass={cs.RO},
      url={https://arxiv.org/abs/2409.01630}, 
}

@misc{zhu2024eairiskbench,
      title={EARBench: Towards Evaluating Physical Risk Awareness for Task Planning of Foundation Model-based Embodied AI Agents}, 
      author={Zihao Zhu and Bingzhe Wu and Zhengyou Zhang and Lei Han and Qingshan Liu and Baoyuan Wu},
      year={2024},
      eprint={2408.04449},
      archivePrefix={arXiv},
      primaryClass={cs.AI},
      url={https://arxiv.org/abs/2408.04449}, 
}

@ARTICLE{liu2024compromising,
  author={Liu, Aishan and Zhou, Yuguang and Liu, Xianglong and Zhang, Tianyuan and Liang, Siyuan and Wang, Jiakai and Pu, Yanjun and Li, Tianlin and Zhang, Junqi and Zhou, Wenbo and Guo, Qing and Tao, Dacheng},
  journal={IEEE Transactions on Information Forensics and Security}, 
  title={Compromising LLM Driven Embodied Agents With Contextual Backdoor Attacks}, 
  year={2025},
  volume={20},
  number={},
  pages={3979-3994},
  doi={10.1109/TIFS.2025.3555410}}

@inproceedings{kim2024openvla,
  author       = {Moo Jin Kim and
                  Karl Pertsch and
                  Siddharth Karamcheti and
                  Ted Xiao and
                  Ashwin Balakrishna and
                  Suraj Nair and
                  Rafael Rafailov and
                  Ethan Paul Foster and
                  Pannag R. Sanketi and
                  Quan Vuong and
                  Thomas Kollar and
                  Benjamin Burchfiel and
                  Russ Tedrake and
                  Dorsa Sadigh and
                  Sergey Levine and
                  Percy Liang and
                  Chelsea Finn},
  editor       = {Pulkit Agrawal and
                  Oliver Kroemer and
                  Wolfram Burgard},
  title        = {OpenVLA: An Open-Source Vision-Language-Action Model},
  booktitle    = {Conference on Robot Learning, 6-9 November 2024, Munich, Germany},
  series       = {Proceedings of Machine Learning Research},
  volume       = {270},
  pages        = {2679--2713},
  publisher    = {{PMLR}},
  year         = {2024},
  url          = {https://proceedings.mlr.press/v270/kim25c.html},
  bibsource    = {dblp computer science bibliography, https://dblp.org}
}

@inproceedings{bharadhwaj2024roboagent,
  author       = {Homanga Bharadhwaj and
                  Jay Vakil and
                  Mohit Sharma and
                  Abhinav Gupta and
                  Shubham Tulsiani and
                  Vikash Kumar},
  title        = {RoboAgent: Generalization and Efficiency in Robot Manipulation via
                  Semantic Augmentations and Action Chunking},
  booktitle    = {{IEEE} International Conference on Robotics and Automation, {ICRA}
                  2024, Yokohama, Japan, May 13-17, 2024},
  pages        = {4788--4795},
  publisher    = {{IEEE}},
  year         = {2024},
  url          = {https://doi.org/10.1109/ICRA57147.2024.10611293},
  doi          = {10.1109/ICRA57147.2024.10611293},
  bibsource    = {dblp computer science bibliography, https://dblp.org}
}

@inproceedings{ma2023eureka,
  author       = {Yecheng Jason Ma and
                  William Liang and
                  Guanzhi Wang and
                  De{-}An Huang and
                  Osbert Bastani and
                  Dinesh Jayaraman and
                  Yuke Zhu and
                  Linxi Fan and
                  Anima Anandkumar},
  title        = {Eureka: Human-Level Reward Design via Coding Large Language Models},
  booktitle    = {The Twelfth International Conference on Learning Representations,
                  {ICLR} 2024, Vienna, Austria, May 7-11, 2024},
  publisher    = {OpenReview.net},
  year         = {2024},
  url          = {https://openreview.net/forum?id=IEduRUO55F},
  bibsource    = {dblp computer science bibliography, https://dblp.org}
}

@article{james2020rlbench,
  title={Rlbench: The robot learning benchmark \& learning environment},
  author={James, Stephen and Ma, Zicong and Arrojo, David Rovick and Davison, Andrew J},
  journal={IEEE Robotics and Automation Letters},
  volume={5},
  number={2},
  pages={3019--3026},
  year={2020},
  publisher={IEEE}
}

@inproceedings{lin2023unlocking,
  author       = {Bill Yuchen Lin and
                  Abhilasha Ravichander and
                  Ximing Lu and
                  Nouha Dziri and
                  Melanie Sclar and
                  Khyathi Raghavi Chandu and
                  Chandra Bhagavatula and
                  Yejin Choi},
  title        = {The Unlocking Spell on Base LLMs: Rethinking Alignment via In-Context
                  Learning},
  booktitle    = {The Twelfth International Conference on Learning Representations,
                  {ICLR} 2024, Vienna, Austria, May 7-11, 2024},
  publisher    = {OpenReview.net},
  year         = {2024},
  url          = {https://openreview.net/forum?id=wxJ0eXwwda},
  bibsource    = {dblp computer science bibliography, https://dblp.org}
}

@online{GPT2024,
author ={OpenAI},
year = {2025},
title ={Models - OpenAI API},
url ={https://platform.openai.com/docs/overview},
month =mar,
lastaccessed ={March 22, 2025},
}

@online{Qwen2024,
author ={Alibaba Cloud},
year = {2025},
title ={Tongyi Qianwen (Qwen) - Alibaba Cloud},
url ={https://www.alibabacloud.com/en/solutions/generative-ai/qwen?_p_lc=1},
month =mar,
lastaccessed ={March 22, 2025},
}

@online{DeepSeek2025,
author ={DeepSeek},
year = {2025},
title ={DeepSeek API Docs},
url ={https://api-docs.deepseek.com/},
month =mar,
lastaccessed ={March 22, 2025},
}

@misc{meng2025embodied,
      title={Embodied Long Horizon Manipulation with Closed-loop Code Generation and Incremental Few-shot Adaptation}, 
      author={Yuan Meng and Xiangtong Yao and Haihui Ye and Yirui Zhou and Shengqiang Zhang and Zhenguo Sun and Xukun Li and Zhenshan Bing and Alois Knoll},
      year={2025},
      eprint={2503.21969},
      archivePrefix={arXiv},
      primaryClass={cs.RO},
      url={https://arxiv.org/abs/2503.21969}, 
}

@misc{yao2023bridging,
      title={Bridging Language and Action: A Survey of Language-Conditioned Robot Manipulation}, 
      author={Xiangtong Yao and Hongkuan Zhou and Oier Mees and Yuan Meng and Ted Xiao and Yonatan Bisk and Jean Oh and Edward Johns and Mohit Shridhar and Dhruv Shah and Jesse Thomason and Kai Huang and Joyce Chai and Zhenshan Bing and Alois Knoll},
      year={2026},
      eprint={2312.10807},
      archivePrefix={arXiv},
      primaryClass={cs.RO},
      url={https://arxiv.org/abs/2312.10807}, 
}

@article{10.1145/3719664,
author = {Zheng, Yue and Chen, Yuhao and Qian, Bin and Shi, Xiufang and Shu, Yuanchao and Chen, Jiming},
title = {A Review on Edge Large Language Models: Design, Execution, and Applications},
year = {2025},
issue_date = {August 2025},
publisher = {Association for Computing Machinery},
address = {New York, NY, USA},
volume = {57},
number = {8},
issn = {0360-0300},
url = {https://doi.org/10.1145/3719664},
doi = {10.1145/3719664},
journal = {ACM Comput. Surv.},
month = mar,
articleno = {209},
numpages = {35}
}

\newpage

\appendix

\section*{Appendix}

\section{Platform}
\label{sec:Platform}
Our experiments in simulation are conducted on a server running a 64-bit Ubuntu 20.04.6LTS system with AMD\textsuperscript{\textregistered} EPYC 7402 @2.8GHz 24-core processors, 128GB memory, and four Nvidia RTX3090 GPUs, each with 24GB memory. The experiments are performed using Python3.9.13. Our experiments on the real-world systems are conducted on a 6-DoF myCobot robot arm with a myCobot gripper from Elephant Robotics. In addition, we employ a compact Femto Bolt camera equipped with multi-modal depth and RGB sensors. It is manufactured by Orbbec.

\section{Details in VoxPoser Framework Experiment}

\subsection{RoboInspector Prompt Template}
\label{sec:VoxPoser-prompt}
We describe the prompts used for each experiment process in our pipeline:

\newtcolorbox{codebox}[1][]{
    colback=gray!10!white,
    colframe=gray!30!black,
    breakable,  
    boxrule=1.5pt,  
    left=5mm,
    right=5mm,
    top=3mm,
    bottom=3mm,
    fonttitle=\bfseries\sffamily,  
    before upper={\parindent0pt}, 
    #1
}

\begin{codebox}[title=Prompt for unveiling the unreliable policy code]
\textbf{"role": "system", "content":}\\
"You are a helpful assistant that pays attention to the user's instructions and writes good python code for operating a robot arm in a tabletop environment."

\bigskip

\textbf{"role": "user", "content":}\\
"I'm seeking assistance with developing Python code to manage a robotic arm functioning on a tabletop. Each time I present a new prompt, please give the policy code accordingly. Focus closely on identifying and maintaining consistent coding structures from the context provided. Ensure your solutions are meticulous and well-considered. Omit any import statements. Avoid restating my requests or adding textual explanations (inline code comments are acceptable). To begin, here’s the reference code segment: `{Demonstration code}`. Note: the coordinate system is defined as follows — x indicates depth (front to back), y represents horizontal movement (left to right), and z denotes vertical direction (bottom to top)."

\bigskip

\textbf{"role": "assistant", "content":}\\
"Got it. I will provide policy code what you give me next."

\bigskip

\textbf{"role": "user", "content":}\\
"\# Query: put rubbish in bin."

\bigskip

\textbf{Example output:}\\
"planner" generated code \\
context: "objects = ['bin', 'rubbish', 'tomato1', 'tomato2']" \\
composer(grasp the rubbish)\\
composer(back to default pose) \\
composer(open gripper)\\
composer(move to the top of the bin) . . .\\
\end{codebox}

\clearpage 
\subsection{Unreliable Behaviors Statistics}
\label{sec:VoxPoser-statistics}

\begin{table}[htbp]
\centering
\caption{\textbf{Detailed unreliable behaviors statistics on 168 distinct combinations of tasks, instruction, and LLM with VoxPoser framework.} \textbf{Ins.} stands for Instructions.}
\label{tab:3}
\resizebox{\textwidth}{!}{%
\begin{tabular}{ccc|cccccccc}
\hline
\multirow{2}{*}{\textbf{Task}} & \multirow{2}{*}{\textbf{Ins.}} & \multirow{2}{*}{\textbf{Unreliable Behaviors}} & \multicolumn{7}{c}{\textbf{Closed-source LLM}} & \textbf{Open-source LLM} \\ \cline{4-11} 
 &  &  & GPT-3.5-turbo & GPT-4 & GPT-4o & GPT-4o-mini & Qwen-max & Qwen-plus & Qwen-turbo & DeepSeek-V3 \\ \hline
\multirow{8}{*}{\textbf{Grasp}} & \multicolumn{1}{c|}{\multirow{4}{*}{${I_A}$}} & \textit{\textbf{Nonsense}} & 9 & 4 & 4 & 2 & 0 & 4 & 24 & 3 \\
 & \multicolumn{1}{c|}{} & \textit{\textbf{Disorder}} & 0 & 0 & 0 & 0 & 0 & 0 & 0 & 0 \\
 & \multicolumn{1}{c|}{} & \textit{\textbf{Infeasible}} & 0 & 0 & 0 & 0 & 0 & 0 & 0 & 0 \\
 & \multicolumn{1}{c|}{} & \textit{\textbf{Badpose}} & 18 & 9 & 11 & 9 & 13 & 8 & 1 & 5 \\ \cline{2-11} 
 & \multicolumn{1}{c|}{\multirow{4}{*}{${I_C}$}} & \textit{\textbf{Nonsense}} & 14 & 0 & 0 & 2 & 0 & 1 & 21 & 0 \\
 & \multicolumn{1}{c|}{} & \textit{\textbf{Disorder}} & 0 & 0 & 0 & 0 & 0 & 0 & 0 & 0 \\
 & \multicolumn{1}{c|}{} & \textit{\textbf{Infeasible}} & 0 & 0 & 0 & 0 & 0 & 0 & 0 & 0 \\
 & \multicolumn{1}{c|}{} & \textit{\textbf{Badpose}} & 3 & 3 & 4 & 5 & 4 & 4 & 0 & 3 \\ \hline
\multirow{8}{*}{\textbf{Movement}} & \multicolumn{1}{c|}{\multirow{4}{*}{${I_A}$}} & \textit{\textbf{Nonsense}} & 14 & 5 & 4 & 0 & 0 & 5 & 21 & 5 \\
 & \multicolumn{1}{c|}{} & \textit{\textbf{Disorder}} & 0 & 0 & 0 & 0 & 0 & 0 & 0 & 0 \\
 & \multicolumn{1}{c|}{} & \textit{\textbf{Infeasible}} & 7 & 5 & 21 & 13 & 11 & 8 & 0 & 5 \\
 & \multicolumn{1}{c|}{} & \textit{\textbf{Badpose}} & 0 & 0 & 0 & 0 & 0 & 0 & 0 & 0 \\ \cline{2-11} 
 & \multicolumn{1}{c|}{\multirow{4}{*}{${I_C}$}} & \textit{\textbf{Nonsense}} & 15 & 0 & 1 & 2 & 0 & 5 & 23 & 0 \\
 & \multicolumn{1}{c|}{} & \textit{\textbf{Disorder}} & 0 & 0 & 0 & 0 & 0 & 0 & 0 & 0 \\
 & \multicolumn{1}{c|}{} & \textit{\textbf{Infeasible}} & 1 & 5 & 4 & 5 & 1 & 2 & 0 & 5 \\
 & \multicolumn{1}{c|}{} & \textit{\textbf{Badpose}} & 0 & 0 & 0 & 0 & 0 & 0 & 0 & 0 \\ \hline
\multirow{8}{*}{\textbf{Rotation}} & \multicolumn{1}{c|}{\multirow{4}{*}{${I_A}$}} & \textit{\textbf{Nonsense}} & 15 & 5 & 3 & 2 & 1 & 1 & 13 & 1 \\
 & \multicolumn{1}{c|}{} & \textit{\textbf{Disorder}} & 0 & 0 & 0 & 0 & 0 & 0 & 0 & 0 \\
 & \multicolumn{1}{c|}{} & \textit{\textbf{Infeasible}} & 10 & 10 & 10 & 10 & 12 & 14 & 10 & 11 \\
 & \multicolumn{1}{c|}{} & \textit{\textbf{Badpose}} & 0 & 0 & 0 & 0 & 0 & 0 & 0 & 0 \\ \cline{2-11} 
 & \multicolumn{1}{c|}{\multirow{4}{*}{${I_C}$}} & \textit{\textbf{Nonsense}} & 18 & 0 & 0 & 3 & 0 & 3 & 20 & 3 \\
 & \multicolumn{1}{c|}{} & \textit{\textbf{Disorder}} & 0 & 0 & 0 & 0 & 0 & 0 & 0 & 0 \\
 & \multicolumn{1}{c|}{} & \textit{\textbf{Infeasible}} & 0 & 5 & 7 & 3 & 5 & 4 & 0 & 4 \\
 & \multicolumn{1}{c|}{} & \textit{\textbf{Badpose}} & 0 & 0 & 0 & 0 & 0 & 0 & 0 & 0 \\ \hline
\multirow{12}{*}{\textbf{SlideBlockToTarget}} & \multicolumn{1}{c|}{\multirow{4}{*}{${I_A}$}} & \textit{\textbf{Nonsense}} & 20 & 0 & 0 & 0 & 6 & 10 & 37 & 1 \\
 & \multicolumn{1}{c|}{} & \textit{\textbf{Disorder}} & 0 & 0 & 0 & 0 & 0 & 0 & 0 & 0 \\
 & \multicolumn{1}{c|}{} & \textit{\textbf{Infeasible}} & 0 & 1 & 4 & 2 & 4 & 10 & 3 & 7 \\
 & \multicolumn{1}{c|}{} & \textit{\textbf{Badpose}} & 16 & 9 & 9 & 15 & 0 & 0 & 0 & 0 \\ \cline{2-11} 
 & \multicolumn{1}{c|}{\multirow{4}{*}{${I_P}$}} & \textit{\textbf{Nonsense}} & 24 & 0 & 0 & 0 & 2 & 0 & 28 & 0 \\
 & \multicolumn{1}{c|}{} & \textit{\textbf{Disorder}} & 0 & 0 & 0 & 0 & 0 & 0 & 0 & 0 \\
 & \multicolumn{1}{c|}{} & \textit{\textbf{Infeasible}} & 0 & 5 & 5 & 6 & 4 & 14 & 3 & 5 \\
 & \multicolumn{1}{c|}{} & \textit{\textbf{Badpose}} & 10 & 1 & 4 & 5 & 0 & 0 & 0 & 2 \\ \cline{2-11} 
 & \multicolumn{1}{c|}{\multirow{4}{*}{${I_C}$}} & \textit{\textbf{Nonsense}} & 24 & 0 & 0 & 0 & 0 & 2 & 28 & 0 \\
 & \multicolumn{1}{c|}{} & \textit{\textbf{Disorder}} & 0 & 0 & 0 & 0 & 0 & 0 & 0 & 0 \\
 & \multicolumn{1}{c|}{} & \textit{\textbf{Infeasible}} & 0 & 5 & 8 & 9 & 3 & 8 & 0 & 3 \\
 & \multicolumn{1}{c|}{} & \textit{\textbf{Badpose}} & 0 & 0 & 0 & 0 & 0 & 0 & 0 & 0 \\ \hline
 \multirow{12}{*}{\textbf{ChangeClock}} & \multicolumn{1}{c|}{\multirow{4}{*}{${I_A}$}} & \textit{\textbf{Nonsense}} & 40 & 4 & 10 & 2 & 0 & 0 & 35 & 4 \\
 & \multicolumn{1}{c|}{} & \textit{\textbf{Disorder}} & 0 & 0 & 0 & 0 & 5 & 7 & 2 & 0 \\
 & \multicolumn{1}{c|}{} & \textit{\textbf{Infeasible}} & 0 & 0 & 0 & 8 & 0 & 0 & 0 & 0 \\
 & \multicolumn{1}{c|}{} & \textit{\textbf{Badpose}} & 0 & 10 & 9 & 5 & 8 & 13 & 0 & 12 \\ \cline{2-11} 
 & \multicolumn{1}{c|}{\multirow{4}{*}{${I_P}$}} & \textit{\textbf{Nonsense}} & 38 & 3 & 4 & 0 & 0 & 0 & 30 & 2 \\
 & \multicolumn{1}{c|}{} & \textit{\textbf{Disorder}} & 0 & 0 & 0 & 0 & 0 & 0 & 0 & 0 \\
 & \multicolumn{1}{c|}{} & \textit{\textbf{Infeasible}} & 0 & 0 & 0 & 8 & 0 & 0 & 0 & 0 \\
 & \multicolumn{1}{c|}{} & \textit{\textbf{Badpose}} & 0 & 8 & 5 & 5 & 9 & 15 & 5 & 10 \\ \cline{2-11} 
 & \multicolumn{1}{c|}{\multirow{4}{*}{${I_C}$}} & \textit{\textbf{Nonsense}} & 28 & 1 & 0 & 2 & 3 & 4 & 30 & 1 \\
 & \multicolumn{1}{c|}{} & \textit{\textbf{Disorder}} & 0 & 0 & 0 & 0 & 0 & 0 & 0 & 0 \\
 & \multicolumn{1}{c|}{} & \textit{\textbf{Infeasible}} & 0 & 1 & 0 & 5 & 0 & 0 & 0 & 2 \\
 & \multicolumn{1}{c|}{} & \textit{\textbf{Badpose}} & 0 & 6 & 6 & 3 & 4 & 8 & 4 & 3 \\ \hline
\multirow{12}{*}{\textbf{LightBulbOut}} & \multicolumn{1}{c|}{\multirow{4}{*}{${I_A}$}} & \textit{\textbf{Nonsense}} & 10 & 0 & 0 & 0 & 0 & 1 & 41 & 0 \\
 & \multicolumn{1}{c|}{} & \textit{\textbf{Disorder}} & 32 & 10 & 7 & 6 & 11 & 14 & 0 & 4 \\
 & \multicolumn{1}{c|}{} & \textit{\textbf{Infeasible}} & 0 & 0 & 0 & 0 & 0 & 0 & 0 & 0 \\
 & \multicolumn{1}{c|}{} & \textit{\textbf{Badpose}} & 0 & 4 & 3 & 9 & 2 & 5 & 0 & 10 \\ \cline{2-11} 
 & \multicolumn{1}{c|}{\multirow{4}{*}{${I_P}$}} & \textit{\textbf{Nonsense}} & 20 & 0 & 0 & 0 & 0 & 0 & 39 & 0 \\
 & \multicolumn{1}{c|}{} & \textit{\textbf{Disorder}} & 1 & 5 & 0 & 2 & 2 & 4 & 0 & 3 \\
 & \multicolumn{1}{c|}{} & \textit{\textbf{Infeasible}} & 0 & 0 & 0 & 0 & 0 & 0 & 0 & 0 \\
 & \multicolumn{1}{c|}{} & \textit{\textbf{Badpose}} & 20 & 6 & 8 & 10 & 8 & 15 & 0 & 10 \\ \cline{2-11} 
 & \multicolumn{1}{c|}{\multirow{4}{*}{${I_C}$}} & \textit{\textbf{Nonsense}} & 24 & 0 & 0 & 0 & 0 & 0 & 37 & 2 \\
 & \multicolumn{1}{c|}{} & \textit{\textbf{Disorder}} & 10 & 2 & 2 & 6 & 8 & 5 & 0 & 2 \\
 & \multicolumn{1}{c|}{} & \textit{\textbf{Infeasible}} & 0 & 0 & 0 & 0 & 0 & 0 & 0 & 0 \\
 & \multicolumn{1}{c|}{} & \textit{\textbf{Badpose}} & 0 & 6 & 3 & 4 & 0 & 6 & 0 & 5 \\ \hline
\end{tabular}%
}
\end{table}

\begin{table}[htbp]
\ContinuedFloat
\centering
\caption{\textbf{(continued)} Detailed unreliable behaviors statistics.}
\resizebox{\textwidth}{!}{%
\begin{tabular}{ccc|cccccccc}
\hline
\multirow{2}{*}{\textbf{Task}} & \multirow{2}{*}{\textbf{Ins.}} & \multirow{2}{*}{\textbf{Unreliable Behaviors}} & \multicolumn{7}{c}{\textbf{Closed-source LLM}} & \textbf{Open-source LLM} \\ \cline{4-11} 
 &  &  & GPT-3.5-turbo & GPT-4 & GPT-4o & GPT-4o-mini & Qwen-max & Qwen-plus & Qwen-turbo & DeepSeek-V3 \\ \hline
\multirow{12}{*}{\textbf{PutRubbishInBin}} & \multicolumn{1}{c|}{\multirow{4}{*}{${I_A}$}} & \textit{\textbf{Nonsense}} & 15 & 0 & 0 & 10 & 2 & 16 & 30 & 1 \\
 & \multicolumn{1}{c|}{} & \textit{\textbf{Disorder}} & 13 & 8 & 12 & 19 & 13 & 7 & 10 & 8 \\
 & \multicolumn{1}{c|}{} & \textit{\textbf{Infeasible}} & 10 & 7 & 6 & 1 & 2 & 4 & 2 & 6 \\
 & \multicolumn{1}{c|}{} & \textit{\textbf{Badpose}} & 0 & 0 & 0 & 0 & 0 & 0 & 0 & 0 \\ \cline{2-11} 
 & \multicolumn{1}{c|}{\multirow{4}{*}{${I_P}$}} & \textit{\textbf{Nonsense}} & 28 & 0 & 2 & 15 & 1 & 12 & 28 & 1 \\
 & \multicolumn{1}{c|}{} & \textit{\textbf{Disorder}} & 5 & 2 & 2 & 5 & 2 & 5 & 2 & 1 \\
 & \multicolumn{1}{c|}{} & \textit{\textbf{Infeasible}} & 2 & 3 & 2 & 5 & 4 & 2 & 2 & 3 \\
 & \multicolumn{1}{c|}{} & \textit{\textbf{Badpose}} & 0 & 3 & 3 & 0 & 7 & 6 & 8 & 5 \\ \cline{2-11} 
 & \multicolumn{1}{c|}{\multirow{4}{*}{${I_C}$}} & \textit{\textbf{Nonsense}} & 28 & 0 & 1 & 15 & 2 & 2 & 33 & 1 \\
 & \multicolumn{1}{c|}{} & \textit{\textbf{Disorder}} & 2 & 2 & 3 & 4 & 4 & 6 & 1 & 2 \\
 & \multicolumn{1}{c|}{} & \textit{\textbf{Infeasible}} & 0 & 2 & 3 & 0 & 5 & 7 & 1 & 2 \\
 & \multicolumn{1}{c|}{} & \textit{\textbf{Badpose}} & 0 & 0 & 0 & 0 & 0 & 0 & 0 & 0 \\ \hline
\multirow{12}{*}{\textbf{OpenWineBottle}} & \multicolumn{1}{c|}{\multirow{4}{*}{${I_A}$}} & \textit{\textbf{Nonsense}} & 35 & 0 & 0 & 0 & 0 & 10 & 47 & 5 \\
 & \multicolumn{1}{c|}{} & \textit{\textbf{Disorder}} & 10 & 10 & 14 & 7 & 0 & 0 & 0 & 0 \\
 & \multicolumn{1}{c|}{} & \textit{\textbf{Infeasible}} & 2 & 10 & 10 & 10 & 7 & 10 & 0 & 15 \\
 & \multicolumn{1}{c|}{} & \textit{\textbf{Badpose}} & 1 & 15 & 20 & 27 & 38 & 15 & 0 & 15 \\ \cline{2-11} 
 & \multicolumn{1}{c|}{\multirow{4}{*}{${I_P}$}} & \textit{\textbf{Nonsense}} & 39 & 0 & 0 & 0 & 3 & 3 & 49 & 0 \\
 & \multicolumn{1}{c|}{} & \textit{\textbf{Disorder}} & 1 & 7 & 2 & 10 & 0 & 0 & 0 & 5 \\
 & \multicolumn{1}{c|}{} & \textit{\textbf{Infeasible}} & 0 & 0 & 9 & 5 & 4 & 0 & 0 & 0 \\
 & \multicolumn{1}{c|}{} & \textit{\textbf{Badpose}} & 8 & 24 & 30 & 30 & 35 & 42 & 0 & 24 \\ \cline{2-11} 
 & \multicolumn{1}{c|}{\multirow{4}{*}{${I_C}$}} & \textit{\textbf{Nonsense}} & 29 & 1 & 0 & 5 & 0 & 2 & 45 & 2 \\
 & \multicolumn{1}{c|}{} & \textit{\textbf{Disorder}} & 1 & 1 & 5 & 1 & 0 & 2 & 0 & 1 \\
 & \multicolumn{1}{c|}{} & \textit{\textbf{Infeasible}} & 1 & 1 & 0 & 5 & 4 & 2 & 0 & 1 \\
 & \multicolumn{1}{c|}{} & \textit{\textbf{Badpose}} & 14 & 22 & 30 & 30 & 30 & 37 & 0 & 20 \\ \hline
\end{tabular}%
}
\end{table}

\subsection{Experiment Scene in VoxPoser Framework}
\label{sec:VoxPoser-scene}

\begin{figure}[H]
  \centering
  \begin{subfigure}{0.18\textwidth}
    \centering
    \includegraphics[width=\linewidth]{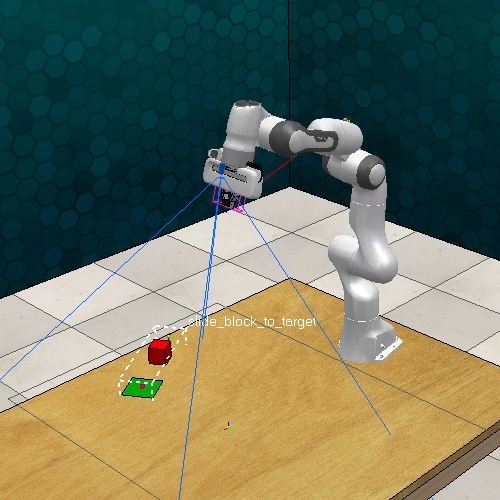}
    \caption{Task \\ SlideBlockToTarget.}
    \label{fig:8-a}
  \end{subfigure}
  \hfill
  \begin{subfigure}{0.18\textwidth}
    \centering
    \includegraphics[width=\linewidth]{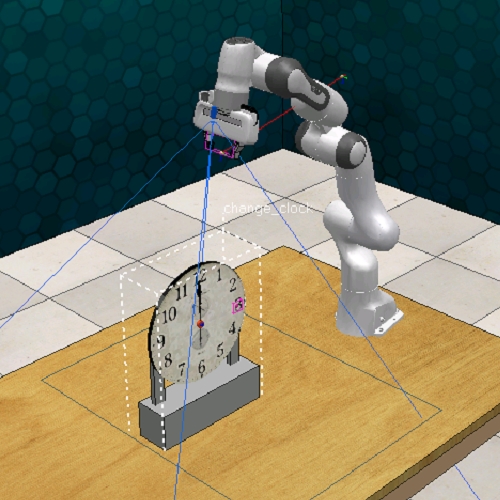}
    \caption{Task \\ ChangeClock.}
    \label{fig:8-b}
  \end{subfigure}
  \hfill
  \begin{subfigure}{0.18\textwidth}
    \centering
    \includegraphics[width=\linewidth]{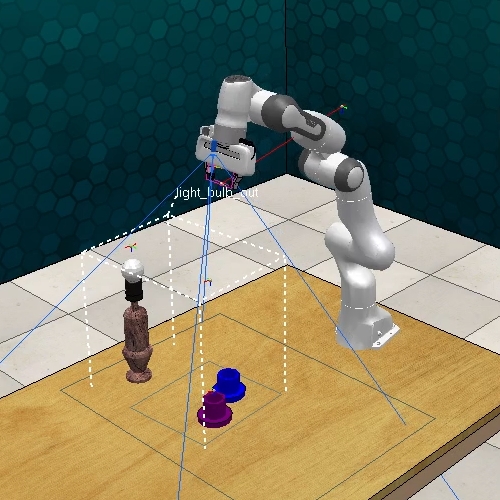}
    \caption{Task \\ LightBulbOut.}
    \label{fig:8-c}
  \end{subfigure}
  \hfill
  \begin{subfigure}{0.18\textwidth}
    \centering
    \includegraphics[width=\linewidth]{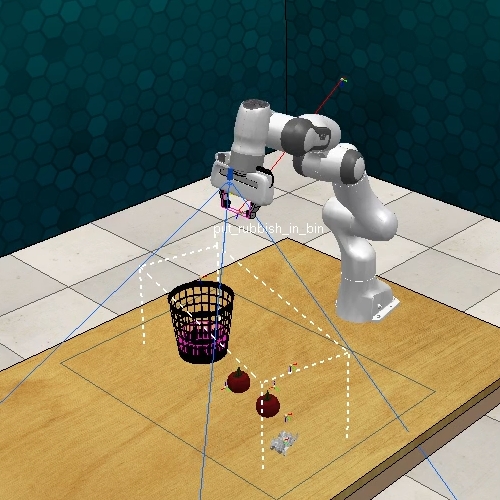}
    \caption{Task \\ PutRubbishInBin.}
    \label{fig:8-d}
  \end{subfigure}
  \hfill
  \begin{subfigure}{0.18\textwidth}
    \centering
    \includegraphics[width=\linewidth]{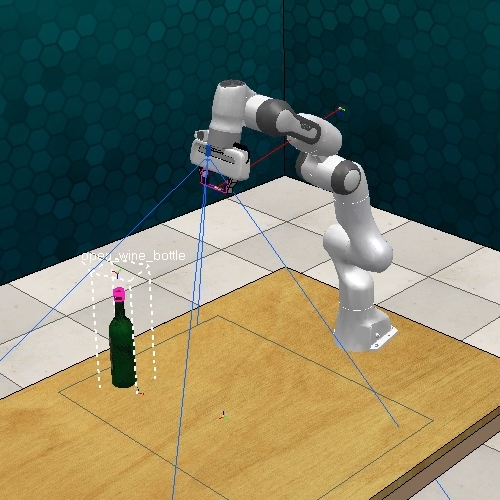}
    \caption{Task \\ OpenWineBottle.}
    \label{fig:8-e}
  \end{subfigure}
  \caption{Experiment scene in VoxPoser framework.}
  \Description{This figure shows the experiment scene in VoxPoser framework.} 
  \label{fig:8}
\end{figure}

\section{Details in Code as Policies Framework Experiment}
\subsection{Experiment Scene in Code as Policies Framework}
\label{sec:CaP-scene}
\begin{figure}[H]
  \centering
  \begin{subfigure}{0.3\textwidth}
    \centering
    \includegraphics[width=\linewidth]{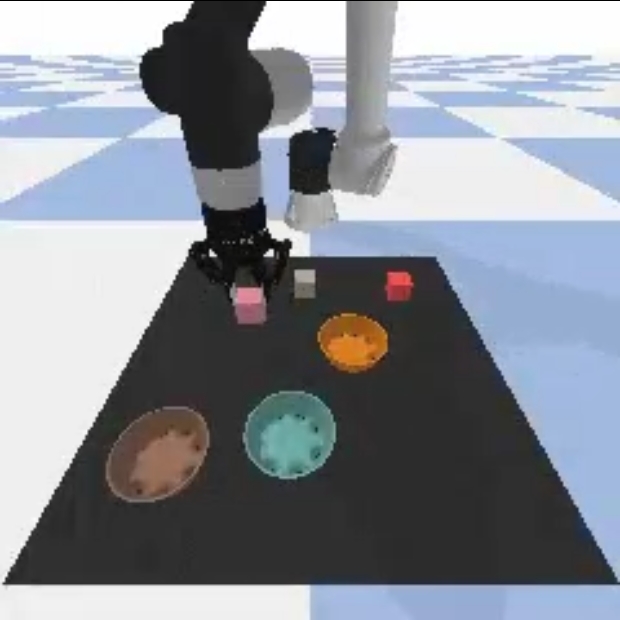}
    \caption{Task Grasp.}
    \label{fig:9-a}
  \end{subfigure}
  \hfill
  \begin{subfigure}{0.3\textwidth}
    \centering
    \includegraphics[width=\linewidth]{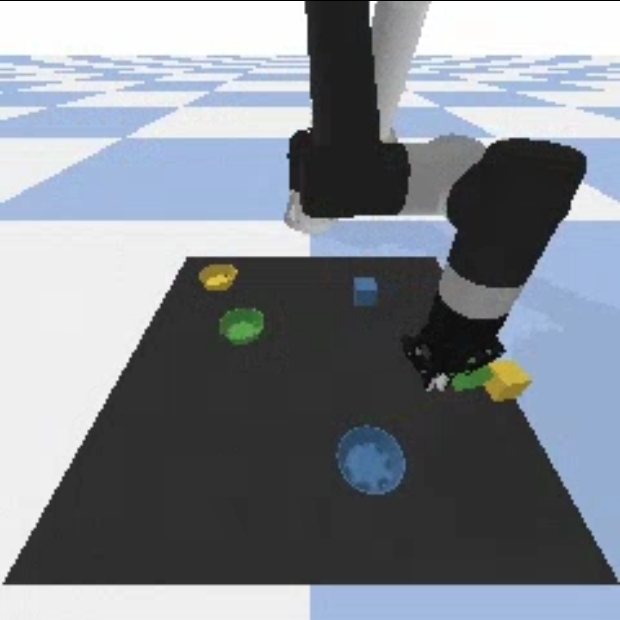}
    \caption{Task Movement.}
    \label{fig:9-b}
  \end{subfigure}
  \hfill
  \begin{subfigure}{0.3\textwidth}
    \centering
    \includegraphics[width=\linewidth]{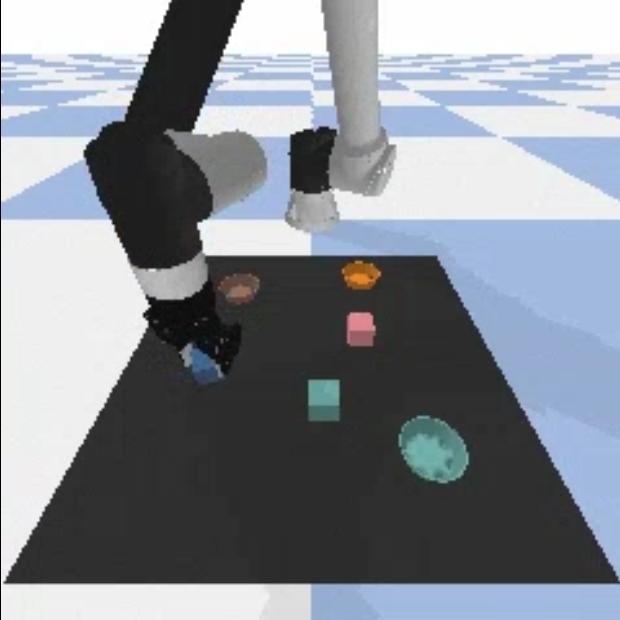}
    \caption{Task Rotation.}
    \label{fig:9-c}
  \end{subfigure}
  \caption{Experiment scene in Code as Policies framework.}
  \Description{This figure shows the experiment scene in Code as Policies framework.} 
  \label{fig:9}
\end{figure}

\subsection{Experimental Results in Code as Policies Framework}
\label{sec:CaP-res}

\begin{table}[H]
\centering
\caption{\textbf{Comparison of success rate on 48 distinct combinations of task, instruction, and LLM with Code as Policies framework.} Black bold underline marks the model with the highest success rate for same instruction, blue bold underline marks the instruction with the highest average success rate for same task. \textbf{Ins.} stands for Instructions.}
\label{tab:4}
\resizebox{\textwidth}{!}{%
\begin{tabular}{@{}cc|ccccccccc@{}}
\toprule
 &  & \multicolumn{7}{c}{\textbf{Closed-source LLM}} & \textbf{Open-source LLM} &  \\ \cmidrule(lr){3-10}
\multirow{-2}{*}{\textbf{Task}} & \multirow{-2}{*}{\textbf{Ins.}} & GPT-3.5-turbo & GPT-4 & GPT-4o & GPT-4o-mini & Qwen-max & Qwen-plus & Qwen-turbo & DeepSeek-V3 & \multirow{-2}{*}{\textbf{\thinspace Avg. \thinspace}} \\ \midrule
 & ${I_A}$ & 0.40 & 0.70 & 0.72 & {\ul \textbf{0.80}} & 0.72 & 0.74 & 0.44 & 0.80 & 0.67 \\
\multirow{-2}{*}{\textbf{Grasp}} & ${I_C}$ & 0.62 & 0.90 & 0.88 & 0.80 & 0.90 & 0.86 & 0.50 & {\ul \textbf{0.92}} & {\color[HTML]{3166FF} {\ul \textbf{0.80}}} \\ \midrule
 & ${I_A}$ & 0.60 & {\ul \textbf{0.90}} & 0.74 & 0.86 & 0.80 & 0.78 & 0.60 & 0.86 & 0.77 \\
\multirow{-2}{*}{\textbf{Movement}} & ${I_C}$ & 0.70 & 0.94 & 0.90 & 0.90 & 0.90 & 0.90 & 0.60 & {\ul \textbf{0.94}} & {\color[HTML]{3166FF} {\ul \textbf{0.85}}} \\ \midrule
 & ${I_A}$ & 0.36 & 0.64 & 0.70 & {\ul \textbf{0.76}} & 0.72 & 0.68 & 0.58 & 0.72 & 0.65 \\
\multirow{-2}{*}{\textbf{Rotation}} & ${I_C}$ & 0.60 & {\ul \textbf{0.92}} & 0.80 & 0.78 & 0.88 & 0.80 & 0.48 & 0.90 & {\color[HTML]{3166FF} {\ul \textbf{0.77}}} \\ \bottomrule
\end{tabular}%
}
\end{table}

\subsection{Statistics on the Proportion of Unreliable Behaviors in Code as Policies Framework}
\label{sec:CaP-Statistics}
\begin{figure}[H]
      \centering
      \includegraphics[width=\linewidth]{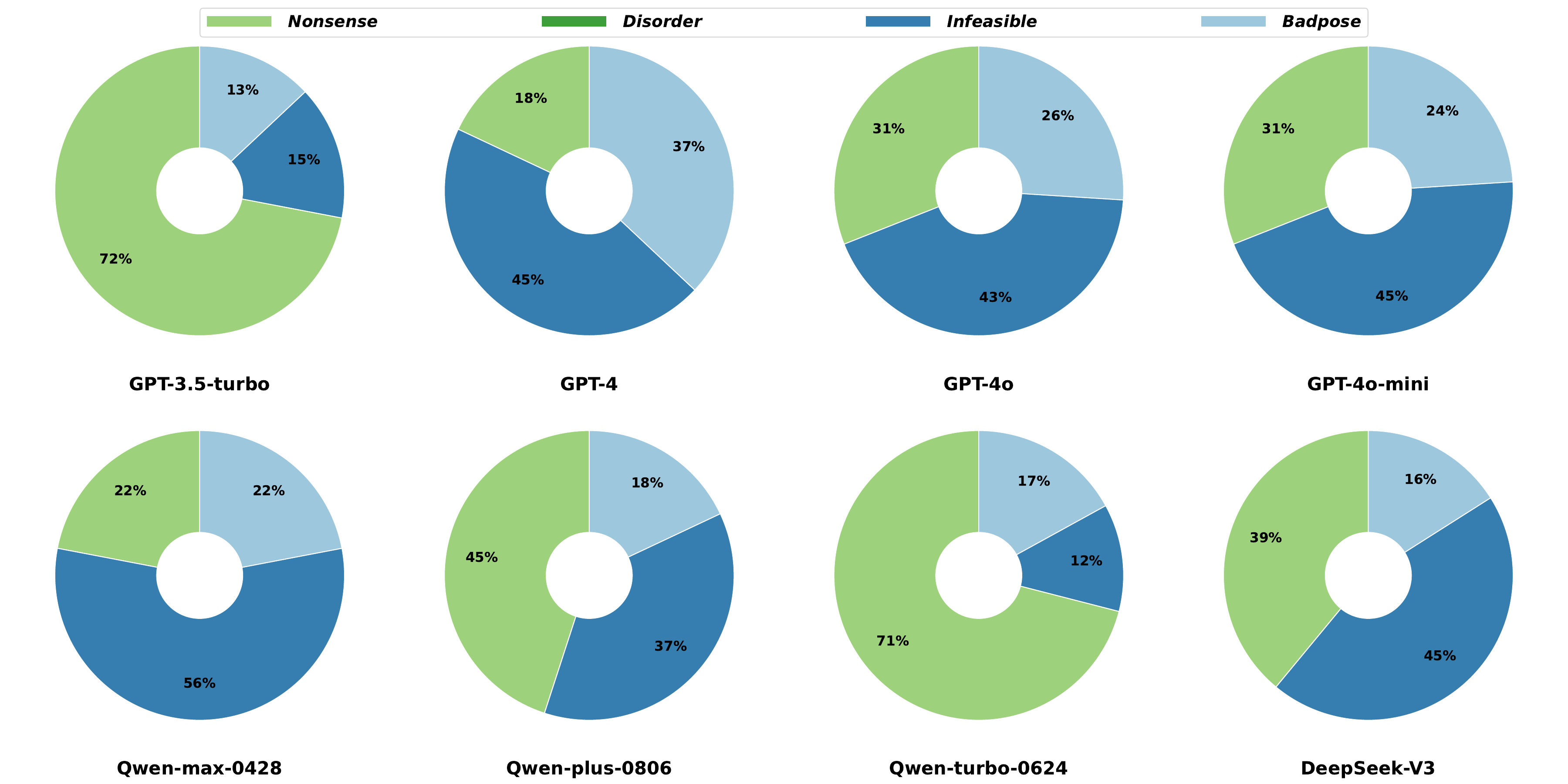}
      \caption{Proportion of unreliable behavior statistics in different LLMs under instruction ${I_A}$.}
      \Description{This figure shows the proportion of unreliable behavior statistics in different LLMs under instruction ${I_A}$.} 
      \label{fig:10}
\end{figure}

\begin{figure}[H]
      \centering
      \includegraphics[width=\linewidth]{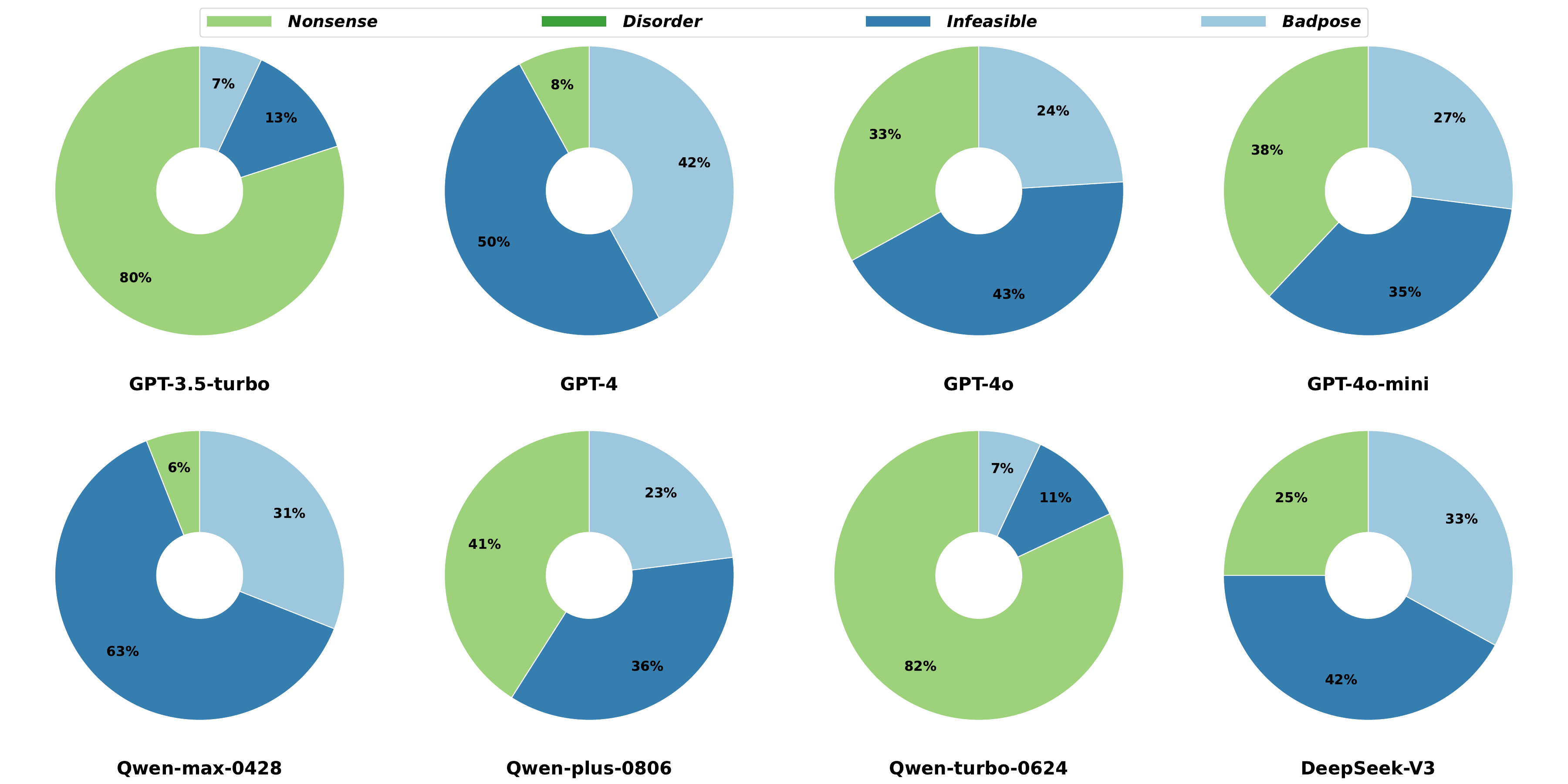}
      \caption{Proportion of unreliable behavior statistics in different LLMs under instruction ${I_C}$.}
      \Description{This figure shows the proportion of unreliable behavior statistics in different LLMs under instruction ${I_C}$.} 
      \label{fig:11}
\end{figure}

\section{Failure Code Feedback Refinement Approach Prompt Template}
\label{refine-prompt}
The following prompt is constructed as a single, self-contained request submitted to the LLM API in one turn. No conversational history or memory is assumed. All information necessary for regeneration, including the original instruction, demonstration code, failed policy code, and behavioral diagnosis is explicitly included within this prompt.

\begin{codebox}[title=Prompt for failure code feedback refinement]
\textbf{"role": "system", "content":}\\
"You are a helpful assistant that pays attention to the user's instructions and writes good python code for operating a robot arm in a tabletop environment."

\bigskip

\textbf{"role": "user", "content":}\\
"During this manipulation, you generated the following failed policy code:\\
import numpy as np\\
$!@()_\#$\\
. . .\\
These policy codes result in one of the following unreliable behaviors:\\
1.Nonsense\\
This behavior refer to the LLM generates policy code that either does not conform to defined criteria or contains irrelevant text.\\
Possible code like: $!@()_\#!@()_\#$\\
Solution is ensure the defined criteria: Omit any import statements. Avoid restating my requests or adding textual explanations.\\
2.Disorder\\
. . .\\
Based on the experience of this failure, regenerate the policy code for the task."

\bigskip

\textbf{Example output:}\\
OK, based on the feedback you provided, regenerated policy code as followed:\\
"planner" generated code \\
context: "objects = ['bin', 'rubbish', 'tomato1', 'tomato2']" \\
composer(grasp the rubbish)\\
composer(back to default pose) \\
composer(move to the top of the bin) \\
composer(open gripper)\\
. . .\\
\end{codebox}

\section{Computational Cost Analysis}
\label{cost}
\autoref{tab:5} reports the token usage and API response latency for each LLM series. Failed manipulation tokens refers to the total token consumption (input + output) of a complete interaction that results in a manipulation failure. Successful manipulation tokens refers to the total token consumption of a complete interaction that results in a successful manipulation. Feedback overhead refers to the additional tokens introduced by the augmented feedback prompt relative to the baseline prompt. Latency is the mean response time per API call, averaged over all tasks under the $I_A$ instruction setting.

\begin{table}[h]
\centering
\caption{Token usage and API call latency under different LLM series. Values are averaged over all tasks and 50 trials per task.}
\label{tab:5}
\resizebox{\textwidth}{!}{%
\begin{tabular}{@{}ccccc@{}}
\toprule
\textbf{Model Series} & \textbf{Failed Manipulation Tokens} & \textbf{Successful Manipulation Tokens} & \textbf{Feedback Overhead Tokens} & \textbf{Latency per Call (s)} \\ \midrule
GPT series & 3,645 & 14,628 & 467 & 3.26 \\ \midrule
Qwen series & 5,475 & 16,848 & 598 & 2.67 \\ \midrule
DeepSeek-V3 & 3,560 & 15,840 & 494 & 6.74 \\ \bottomrule
\end{tabular}%
}
\end{table}


\end{document}